\documentclass[preprint,review,12pt]{elsarticle}
\usepackage{epstopdf}
\makeatletter
\def\ps@pprintTitle{%
	\let\@oddhead\@empty
	\let\@evenhead\@empty
	\let\@oddfoot\@empty
	\let\@evenfoot\@oddfoot
}
\let\subparagraph\relax
\usepackage{amsmath,amssymb,amsbsy,amsfonts,latexsym,amsopn,amstext,amsxtra,euscript,amscd,hyperref}
\usepackage{mathptmx}      
\usepackage{graphicx}
\usepackage{algorithm}
\usepackage{algorithmic}
\usepackage[numbers]{natbib}
\usepackage{notoccite}
\usepackage{graphics,graphicx}
\usepackage{array,tikz}
\usepackage{caption}
\usepackage{slashbox}
\usepackage{wrapfig}
\usepackage{setspace}
\usepackage{titlesec}
\usepackage{enumerate}
\newcolumntype{C}{@{}c@{}}

\usepackage{float}
\usepackage[top=1.5cm, left=1.5cm,right=1.6cm,bottom=1.28cm]{geometry}
\newcounter{myeqno}
\usepackage{chngcntr}

\usetikzlibrary{arrows.meta, positioning, calc}
\usepackage{tikz}
\usetikzlibrary{shapes.geometric, arrows, positioning, fit}
\usetikzlibrary{positioning, shadows, arrows.meta}
\usetikzlibrary{shapes.geometric,arrows,fit,matrix,positioning,shapes.multipart}
\tikzstyle{startstop0}=[rectangle, rounded corners, minimum width=3cm, minimum height=1cm, draw=black]
\tikzstyle{startstop1}=[rectangle, rounded corners, minimum width=8cm, minimum height=4cm, draw=black]
\tikzstyle{startstop2}=[square, rounded corners, minimum width=0.5cm, minimum height=1cm, draw=black]
\tikzstyle{startstop3}=[square, rounded corners, minimum width=5cm, minimum height=1cm, draw=black]
\tikzstyle{startstop4}=[square, rounded corners, minimum width=1cm, minimum height=5cm, draw=black]
\tikzstyle{round} = [ellipse, minimum width=3cm, minimum height=1cm, draw= black]
\tikzstyle{startstop2}=[rectangle, rounded corners, minimum width=2cm, minimum height=1cm, draw=black]
\tikzstyle{arrow} =[draw, -latex']

\tikzstyle{startstop} = [rectangle, rounded corners, minimum width=4cm, minimum height=1cm,text centered, draw=black, fill=green!30]
\tikzstyle{process} = [rectangle, minimum width=3.5cm, minimum height=1cm, text centered, draw=black, fill=blue!20]
\tikzstyle{decision} = [rectangle, minimum width=3.5cm, minimum height=1cm, text centered, draw=black, fill=yellow!30]
\tikzstyle{code} = [rectangle, minimum width=3.5cm, minimum height=1cm, text centered, draw=black, fill=red!30]
\tikzstyle{arrow} = [thick, ->, >=stealth]
\tikzset{
	block/.style={draw, fill=white, rectangle, 
		minimum height=3em, minimum width=6em},
	every path/.style={very thick},}

\usepackage{longtable,lscape}
\usepackage{rotating}
\usepackage{multirow,multicol}
\usepackage{makecell}

\usepackage{stfloats}

\usepackage{rotating}
\usepackage{pdflscape}
\usepackage{xcoffins}
\NewCoffin\tablecoffin

\usepackage{caption,subcaption}
\usepackage{colortbl}
\graphicspath{{images/}}
\definecolor{Burgundy}{RGB}{144,0,32}
\usepackage{newunicodechar}
\newunicodechar{−}{\ensuremath{-}}
\usepackage{comment}
\usepackage{epsfig,epstopdf}

\usepackage{amssymb}
\usepackage{pifont}
\usepackage{booktabs}

\usepackage{appendix}
\usepackage{color,soul}
\usepackage{xcolor}
\usepackage{colortbl}
\definecolor{mygrey}{RGB}{220,220,220}
\usepackage{multirow}
\usepackage{listings}
\newdefinition{example}{Example}
\newdefinition{definition}{Definition}
\newdefinition{remark}{Remark}[section]
\newproof{proof}{Proof}
\newdefinition{theorem}{Theorem}
\begin{document}
\begin{frontmatter}
    \title{Large Language Models and Evolutionary Computation: A Critical Review of Bidirectional Interaction, Automated Algorithm Design, and Co-Adaptive Systems}
    \author[1]{Dikshit Chauhan}
\ead{dikshitchauhan608@gmail.com}
\author[2]{Bapi Dutta }
\ead{bdutta@ujaen.es}
\author[3]{Indu Bala}
\ead{indu.bala@adelaide.edu.au}
\author[4]{Niki Van Stein}
\ead{n.van.stein@liacs.leidenuniv.nl}
\author[4]{Thomas B\"{a}ck}
\ead{t.h.w.baeck@liacs.leidenuniv.nl}
\author[5]{Anupam Yadav \corref{}}
\ead{anupam@nitj.ac.in}
\cortext[]{Corresponding author}

\address[1]{Department of Electrical and Computer Engineering, National University of Singapore, Singapore 119077}
\address[2]{Department of Computer Science, Universidad de Jaén, Spain}
\address[3]{School of Computer and Mathematical Sciences, University of Adelaide, Australia- 5005}
\address[4]{Leiden Institute of Advanced Computer Science, University of Leiden, Netherlands}
\address[5]{Department of Mathematics and Computing, Dr. B. R. Ambedkar National Institute of Technology, Jalandhar, India}

        
\begin{abstract}
Large Language Models (LLMs) and Evolutionary Computation (EC) are increasingly being combined to support automated optimization, algorithm design, and adaptive decision-making. This survey reviews the bidirectional interaction between these two paradigms and examines how their complementary strengths can be leveraged in hybrid intelligent systems. First, we analyze how EC can enhance LLM-based systems through prompt optimization, hyperparameter tuning, and architecture search. Second, we review how LLMs can improve EC by supporting metaheuristic design, surrogate reasoning, adaptive operator control, and heuristic generation. We further discuss emerging co-adaptive frameworks in which LLMs and EC interact through iterative feedback loops. Beyond summarizing recent developments, the survey provides a structured perspective on interaction mechanisms, application patterns, and methodological challenges, including computational cost, reproducibility, interpretability, benchmarking, and generalization. The paper concludes by outlining open research questions and future directions for developing more robust, transparent, and scalable LLM-EC systems.
\end{abstract}
\end{frontmatter}

\section{Introduction}
Large Language Models (LLMs) have become a central paradigm in artificial intelligence (AI), demonstrating strong capabilities in natural language understanding, generation, reasoning, and code synthesis \cite{zhao2023survey,minaee2025largelanguagemodelssurvey}. Built primarily on Transformer-based architectures and trained on massive corpora, these models can perform a wide range of tasks with high fluency and adaptability, often under zero-shot and few-shot settings. Their growing impact now extends well beyond traditional natural language processing (NLP), influencing software engineering, scientific discovery, decision support, and increasingly, automated optimization and algorithm design. As a result, LLMs are no longer viewed only as language interfaces but also as computational components that can guide search, generate heuristics, and support the automation of complex design workflows.

In parallel, Evolutionary Computation (EC) has long served as a powerful family of derivative-free optimization methods inspired by natural evolution \cite{bartz2014evolutionary,slowik2020evolutionary}. By iteratively evolving populations of candidate solutions through variation and selection, EC methods are well-suited to high-dimensional, multimodal, non-convex, and non-differentiable search spaces where classical optimization techniques may struggle. Their robustness, flexibility, and model-agnostic nature have made them valuable across engineering design, machine learning, scheduling, scientific modeling, and many other domains. These same properties also make EC particularly attractive for improving or steering LLM-based systems, especially in black-box settings where gradients, internals, or exact performance landscapes may be inaccessible.

From a broader computer-science perspective, the intersection of LLMs and EC is important not only because it may improve performance, but because it speaks directly to a larger shift toward \emph{automated algorithm design}. LLMs contribute semantic abstraction, prior knowledge, code generation, and flexible reasoning, whereas EC contributes structured exploration, diversity preservation, black-box optimization, and robustness under weak modeling assumptions. Their combination, therefore, offers a promising route toward AI systems that can not only solve tasks but also help design, adapt, and optimize the procedures used to solve them.

\subsection{Overview of LLMs and EC}\label{subsec:formal_definitions}

\begin{definition}{Large Language Model (LLM):}
An LLM is a deep neural network that models the conditional probability distribution of a token sequence $T=\{t_1, t_2, \ldots, t_n\}$ as
\begin{equation*}
    p(t_n \mid t_1, t_2, \ldots, t_{n-1}; \theta),
\end{equation*}
where $\theta$ denotes the model parameters learned through large-scale self-supervised training, typically via next-token prediction. In practice, modern LLMs leverage Transformer-based architectures, large pretraining corpora, and transfer learning to acquire contextual, semantic, and task-general reasoning capabilities.
\end{definition}

\begin{definition}{Evolutionary Computation (EC):}
EC refers to a family of population-based stochastic optimization methods inspired by natural evolution. An EC algorithm evolves a population of candidate solutions $P_t=\{x_1, x_2, \ldots, x_N\}$ over generations using variation and selection operators to optimize an objective function $f(x)$:
\begin{equation*}
    x^{*}=\arg \max_{x \in X} f(x),
\end{equation*}
or equivalently $\arg \min$ for minimization problems. Representative EC paradigms include genetic algorithms (GA), differential evolution (DE), evolution strategies (ES), genetic programming (GP), and swarm-based methods. Their distinguishing characteristics include diversity-driven exploration, robustness in complex search spaces, and applicability in settings where gradients are noisy, unavailable, or uninformative \cite{CHAUHAN2025102048}.
\end{definition}
{\color{black}In this review, we use the term \emph{evolutionary computation} (EC) in a broad computer-science sense. Specifically, EC serves here as an umbrella term for population-based and evolution-inspired search methods, including canonical evolutionary algorithms such as genetic algorithms, differential evolution, evolution strategies, and genetic programming, as well as closely related population-based metaheuristics such as swarm-intelligence methods when they play a methodologically comparable role in LLM-related optimization and algorithm-design settings. By contrast, \emph{evolutionary algorithms} (EAs) are treated as a more specific subset within this broader landscape, while \emph{metaheuristics} is used as a higher-level term for general-purpose heuristic optimization frameworks that may include both evolutionary and non-evolutionary methods \cite{blum2003meta}. For convenience, Table~\ref{tab:terminology} summarizes the main terminology and abbreviations used throughout the survey.
}
{\color{black}\begingroup
\footnotesize\begin{longtable}{p{5.5cm}p{1cm}p{11cm}}
\caption{Important terminology and abbreviations used throughout the survey.}
\label{tab:terminology}\\

\toprule
\textbf{Term} &   & \textbf{Definition / Description} \\
\midrule
\endfirsthead

\multicolumn{3}{c}{\tablename\ \thetable\ - Continued from previous page}\\
\toprule
\textbf{Term} &  & \textbf{Definition / Description} \\
\midrule
\endhead

\bottomrule
\endfoot

Large Language Model & LLM & Transformer-based deep neural network trained on massive text corpora for language understanding, reasoning, and generation. \\

Evolutionary Computation & EC & Population-based optimization paradigm inspired by biological evolution. \\

Evolutionary Algorithm & EA & Subclass of EC employing selection, crossover, mutation, and replacement operators. \\

 Genetic Algorithm & GA & Evolutionary optimization algorithm based on chromosome evolution. \\

 Differential Evolution & DE & Population-based optimization algorithm using vector differences for mutation. \\

Evolution Strategies & ES & Evolutionary optimization methods emphasizing adaptive mutation mechanisms. \\

Genetic Programming & GP & Evolutionary technique that evolves executable programs or symbolic expressions. \\

Neural Architecture Search & NAS & Automated optimization of neural network architectures. \\


Prompt Optimization & PO & Automatic optimization of prompts to improve LLM performance. \\



Hyper-Heuristic & HH & Method that searches over heuristics. \\

Metaheuristic & MH & General-purpose optimization framework applicable to diverse problems. \\

Surrogate Model & SM & Approximate model used in place of expensive objective evaluations. \\

Fitness Function & FF & Objective function used to evaluate candidate solutions. \\

Population & -- & Collection of candidate solutions evolved during optimization. \\

Individual & -- & Single candidate solution within a population. \\

Selection & -- & Process of selecting promising individuals for reproduction. \\

Mutation & -- & Random modification of candidate solutions to introduce diversity. \\

Crossover & -- & Recombination of parent solutions to generate offspring. \\

Exploration & -- & Search strategy emphasizing diversity and global search. \\

Exploitation & -- & Search strategy emphasizing refinement of promising solutions. \\

Quality-Diversity & QD & Optimization paradigm for high-quality and diverse solutions simultaneously. \\




Particle Swarm Optimization & PSO & Swarm intelligence algorithm inspired by flocking behavior. \\

Artificial Bee Colony & ABC & Swarm optimization algorithm inspired by honeybee foraging. \\

Grey Wolf Optimizer & GWO & Nature-inspired optimization algorithm based on grey wolf hunting behavior. \\

Whale Optimization Algorithm & WOA & Population-based optimizer inspired by humpback whale hunting. \\

Self-Organizing Migrating Algorithm & SOMA & Evolutionary optimization algorithm based on cooperative migration. \\

Ant Colony Optimization & ACO & Optimization algorithm inspired by pheromone-guided ant foraging. \\

Bayesian Optimization & BO & Sample-efficient optimization for expensive black-box functions. \\


Black-box Optimization & BBO & Optimization without explicit knowledge of gradients or objective structure. \\


Prompt Engineering & PE & Manual or automated design of prompts for LLMs. \\

Automated Algorithm Design & AAD & Automatic generation and optimization of algorithms using AI techniques. \\

Program Synthesis & PS & Automatic generation of executable programs satisfying desired objectives. \\

Algorithm Discovery & AD & Automatic creation of novel optimization algorithms. \\





FunSearch & -- & LLM-based framework integrating code generation with evolutionary evaluation. \\

LLaMEA & -- & LLM-assisted evolutionary algorithm generation and evaluation framework. \\




Co-Adaptive System & -- & Hybrid framework where LLMs and EC iteratively improve one another through feedback. \\

Co-Evolutionary System & -- & Framework in which both LLMs and EC undergo mutual evolutionary adaptation. \\


Search Space & -- & Domain containing all feasible candidate solutions. \\

Objective Function & -- & Mathematical function defining the optimization objective. \\

Benchmark & -- & Standard dataset or optimization problem used for evaluation. \\



Explainability & XAI & Ability to interpret and understand model predictions and optimization behavior. \\

\end{longtable}\endgroup}

These definitions reveal a useful complementarity. LLMs excel at representing and generating structured knowledge, while EC excels at searching difficult spaces under limited structural assumptions. Recent advances in LLM capability have therefore intensified interest in using EC to optimize prompts, hyperparameters, architectures, and interaction policies of LLM-based systems. Conversely, LLMs are increasingly being used to assist EC by generating heuristics, proposing operators, providing surrogate judgments, and supporting automated metaheuristic design. This bidirectional interaction motivates a more unified view of how the two paradigms exchange information, constraints, and search bias.

The convergence of LLMs and EC thus creates opportunities at multiple levels of abstraction. In one direction, EC can improve LLM-centered pipelines through prompt optimization, hyperparameter tuning, neural architecture search, and other black-box adaptation mechanisms \cite{guo2024connectinglargelanguagemodels,akiba2025evolutionary,nasir2024llmatic,yin2021autotinybertautomatichyperparameteroptimization}. In the opposite direction, LLMs can improve EC by generating promising candidate solutions, proposing variation operators, designing heuristics, and injecting domain knowledge into evolutionary search \cite{pluhacek2024using,pluhacek2023leveraging,liu2024large,hao2024large}. { \color{black}Beyond these two direct modes, a third emerging direction involves \emph{co-adaptive} or \emph{co-evolutionary} frameworks in which LLMs and EC iteratively influence one another through feedback loops. In this review, \emph{co-adaptive} is used as the broader term for repeated mutual adjustment between the two components, whereas \emph{co-evolutionary} is reserved for the narrower case in which both components explicitly undergo coupled adaptation or evolutionary update in response to each other.}
Fig.~\ref{fig:llm_ec_timeline} summarizes the evolution of LLM-EC interaction from foundational language-model advances to prompt optimization, automated algorithm design, co-adaptive systems, and the recent emphasis on benchmarking and reproducibility. {\color{black}The timeline is intended as a synthesis of representative developments discussed later in the paper rather than as an exhaustive publication chronology.} 

\begin{figure}[h!]
\centering
\begin{tikzpicture}[
    >=Stealth,
    timeline/.style={thick, -{Stealth[length=3mm]}},
    milestone/.style={circle, draw, fill=white, minimum size=4.5mm, inner sep=0pt},
    box/.style={draw, rounded corners, align=center, text width=3.6cm, font=\small, inner sep=5pt}
]
\draw[timeline, line width=2pt] (0,0) -- (14.5,0);
\foreach \x/\yr in {
    1.0/{2017-2020},
    3.1/{2022},
    5.2/{2023},
    7.3/{2024},
    9.4/{2024-2025},
    11.5/{2025},
    13.6/{2025-2026}
}{
    \node[milestone] at (\x,0) {};
    \node[font=\scriptsize, below=4pt] at (\x,0) {\yr};
}

\node[box, above=1.3cm, fill=orange!10] (b1) at (1.0,0)
{Transformer/GPT-era foundation};

\node[box, above=1.3cm, fill=orange!20] (b3) at (5.2,0)
{LLM-generated operators and hybrid metaheuristics};

\node[box, above=1.3cm, fill=orange!30] (b5) at (9.4,0)
{Language-mediated evolutionary search};

\node[box, above=1.3cm, fill=orange!40] (b7) at (13.6,0)
{Benchmarking and reproducibility phase};

\node[box, below=1.35cm, fill=blue!10] (b2) at (3.1,0)
{Prompt optimization with EC};

\node[box, below=1.35cm, fill=blue!20] (b4) at (7.3,0)
{Hyper-heuristic/program-synthesis frameworks};

\node[box, below=1.35cm, fill=blue!30] (b6) at (11.5,0)
{Co-adaptive LLM-EC systems};

\draw[-] (1.0,0.2) -- (b1.south);
\draw[-] (5.2,0.2) -- (b3.south);
\draw[-] (9.4,0.2) -- (b5.south);
\draw[-] (13.6,0.2) -- (b7.south);

\draw[-] (3.1,-0.2) -- (b2.north);
\draw[-] (7.3,-0.2) -- (b4.north);
\draw[-] (11.5,-0.2) -- (b6.north);
\end{tikzpicture}
\caption{\color{black}Evolution of LLM-EC interaction, from foundational LLM advances to prompt optimization, algorithm-generation frameworks, language-mediated search, co-adaptive systems, and the growing emphasis on benchmarking and reproducibility. Representative milestones reflected in the timeline include foundational LLM surveys and model developments \cite{zhao2023survey,minaee2025largelanguagemodelssurvey}, early evolutionary prompt optimization studies \cite{xu2022gps,guo2024connectinglargelanguagemodels}, language-mediated evolutionary search \cite{liu2024large}, and benchmarking-oriented algorithm-generation frameworks such as LLaMEA \cite{van2025llamea}.}
\label{fig:llm_ec_timeline}
\end{figure}

{\color{black}To clarify this relationship, Table~\ref{tab:comparison1} provides a compact comparison of the complementary properties of LLMs and EC across learning, search, and optimization dimensions. Rather than treating them as competing paradigms, the table highlights why they are natural partners in automated design and optimization workflows and serves as a conceptual backdrop for the bidirectional interaction patterns discussed later in the paper.}

\begin{table}[h!]
\centering
\caption{\color{black}Compact comparison of LLMs and EC across learning, search, and optimization dimensions. The table highlights their complementary properties and provides a conceptual backdrop for the bidirectional interaction patterns discussed later in the survey.}
\label{tab:comparison1}
\resizebox{1\linewidth}{!}{
\begin{tabular}{p{3.6cm}p{10.0cm}p{9.0cm}}
\toprule
\textbf{Dimension} & \textbf{Large Language Models (LLMs)} & \textbf{Evolutionary Computation (EC)} \\
\midrule
Learning Properties
&
\textit{Core principle:} Learn statistical and semantic structure from large-scale token data. \newline
\textit{Learning mechanism:} Gradient-based representation learning from pretraining and adaptation data. \newline
\textit{Knowledge representation:} Distributed latent embeddings and contextual token representations.
&
\textit{Core principle:} Evolve candidate solutions through variation, selection, and population adaptation. \newline
\textit{Learning mechanism:} Fitness-driven stochastic search over populations of solutions. \newline
\textit{Knowledge representation:} Explicit symbolic, numeric, or structural encodings of candidate solutions.
\\
\midrule
Search Properties
&
High-dimensional semantic and latent spaces, with exploration and exploitation shaped indirectly through prompting, decoding, and learned priors from pretraining and instruction tuning.
&
Problem-dependent decision spaces over parameters, rules, structures, or programs, with exploration and exploitation controlled explicitly through mutation, recombination, selection pressure, diversity maintenance, and fitness design.
\\
\midrule
Optimization Properties
&
Objectives typically involve likelihood, utility, alignment, or downstream task quality. Main strengths include rich semantic reasoning, cross-task transfer, and natural-language/code-level abstraction. Main weaknesses include hallucination, bias, prompt sensitivity, high compute demand, and limited transparency of internal reasoning.
&
Objectives are defined through user-specified fitness functions. Main strengths include robust global search in black-box, non-differentiable, and multimodal spaces. Main weaknesses include sensitivity to representation and fitness design, risk of premature convergence, and high evaluation cost for expensive objectives.
\\
\bottomrule
\end{tabular}}
\end{table}

\begin{figure}[htbp]
    \centering
    \includegraphics[width=0.7\linewidth]{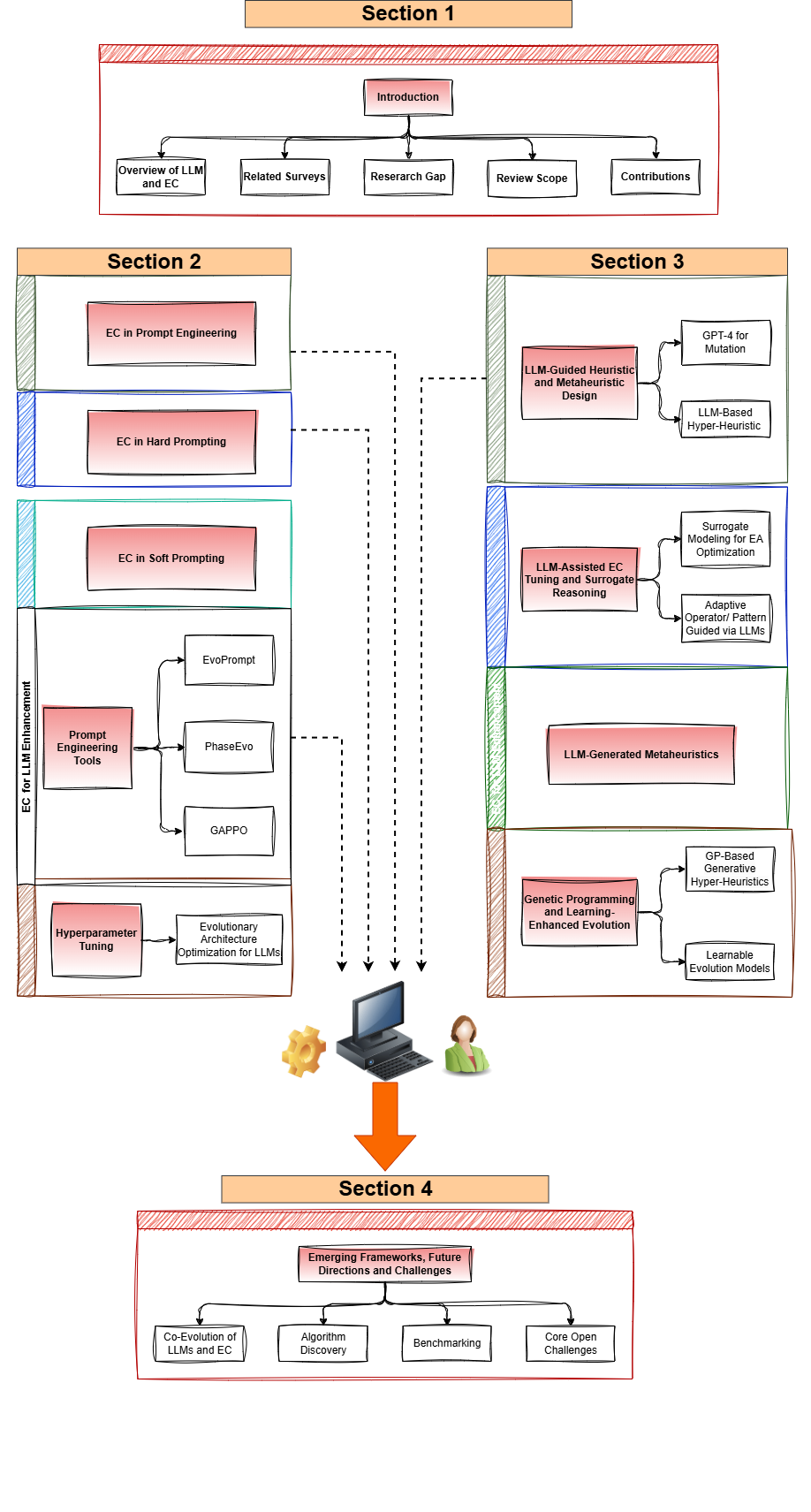}
    \caption{\color{black}Overview of the survey structure around three complementary perspectives: EC for LLM enhancement, LLMs for EC enhancement, and emerging co-adaptive frameworks that connect the two through iterative feedback and refinement.}
    \label{fig:intro}
\end{figure}

Among the most visible examples of this integration is prompt optimization. Since LLM outputs are highly sensitive to prompt wording and structure, prompt engineering has become an important lever for improving downstream performance. However, manually identifying high-quality prompts is labor-intensive and often task-specific. Evolutionary prompt optimization frameworks such as \texttt{EvoPrompt} automate this search by generating and selecting improved prompt variants over time \cite{guo2024connectinglargelanguagemodels,saletta2024exploring}. More broadly, prompt optimization illustrates a recurring pattern throughout this survey: EC provides efficient search over difficult design spaces, while LLMs provide semantic structure and evaluative priors that can reshape those searches.

{\color{black}More generally, hybrid LLM-EC systems are most useful when neither component alone is sufficient for the design problem at hand. In particular, their combination is especially justified when the object being optimized has semantic, textual, or code-level structure, but its quality must still be assessed through a black-box, non-differentiable, multimodal, expensive, or weakly structured evaluation process. This situation arises, for example, when prompts, heuristics, candidate algorithms, or interaction policies must be generated or refined using contextual knowledge, while robust global exploration and iterative performance feedback remain necessary. More broadly, the practical value of LLM-EC integration is strongest when manual design is costly, gradient-based adaptation is unavailable or unreliable, and purely language-based generation is not sufficient to ensure systematic search, robustness, or reproducible improvement.

 \color{black}The broader significance of this interaction lies not only in improved performance, but in its potential to support automated design workflows in which semantic generation and evolutionary search complement one another. As illustrated in Fig.~\ref{fig:intro}, this survey treats these interactions as a structured design space organized around EC for LLM enhancement, LLMs for EC enhancement, and emerging co-adaptive systems.}


\subsection{Related Surveys and Research Gaps}
The literature surrounding LLMs and EC has expanded rapidly, and several recent surveys now cover adjacent parts of this space. Consequently, the key research gap is no longer the absence of prior reviews, but the absence of a \emph{balanced, EC-aware, and critically synthesized} account of how LLMs and EC interact across representation, search, automation, evaluation, and application levels.

Existing survey literature can be grouped into four broad categories. First, general LLM surveys established the architectural, training, and application foundations of large language models, but did not focus on optimization or evolutionary computation \cite{zhao2023survey}. Second, optimization-oriented surveys examined LLMs as optimizers, problem solvers, or assistants for algorithmic tasks, thereby clarifying the LLM$\rightarrow$optimization direction but offering limited treatment of EC-specific mechanisms \cite{yu2024deep}. Third, recent reviews have explicitly addressed the interaction between EC and LLMs, most notably through a survey-and-roadmap perspective centered on evolutionary computation in the era of LLMs \cite{wu2025evolutionary,wang2025large}. Fourth, broader reviews on LLMs for algorithm design and LLM-driven meta-optimizers have further expanded the landscape by covering automated algorithm design, meta-optimization, and heuristic generation under a wider methodological umbrella \cite{liu2026systematic,zheng2026survey}.

These works have significantly advanced the field, but they also reveal a remaining need for a survey tailored to the goals of this paper. Existing reviews tend to emphasize one of the following: (i) general algorithm design with LLMs, (ii) EC research in the LLM era from an EA-centered viewpoint, or (iii) LLM-driven meta-optimization at a broader level that is not limited to EC. What remains insufficiently synthesized is a unified treatment of \emph{EC for LLM enhancement}, \emph{LLM for EC enhancement}, and \emph{co-adaptive LLM-EC systems}, together with a critical discussion of benchmarking practice, reproducibility, generalization, computational cost, and deployment considerations. Table~\ref{tab:comparison} summarizes the most relevant review papers and closely related survey-style works, highlighting their scope, key contribution, and the specific gap they leave open for the present survey.

\begin{table}[h!]
\centering
\caption{Chronological summary and comparative analysis of the most relevant prior surveys and review-oriented works related to LLM-EC integration.}
\label{tab:comparison}
\resizebox{1\linewidth}{!}{\begin{tabular}{p{1.6cm} p{2.7cm} p{2.5cm} p{5.6cm} p{8.4cm}}
\toprule
\textbf{Year} & \textbf{Reference} & \textbf{Scope} & \textbf{Key Contribution} & \textbf{Main Limitation Relative to This Survey} \\
\midrule

2023 & Zhao et al.~\cite{zhao2023survey} & General LLMs & Comprehensive survey of the background, key findings, and mainstream techniques of LLMs, with emphasis on pre-training, adaptation tuning, utilization, and capacity evaluation. & Does not focus on optimization, evolutionary computation, or the bidirectional interaction between LLMs and EC. \\
2024 & Yu et al.~\cite{yu2024deep} & Optimization with AI & Reviews automated optimization with deep generative models and LLMs. & Treats EC only partially and does not develop a dedicated LLM-EC taxonomy. \\

2025 & Wu et al.~\cite{wu2025evolutionary} & EC $\leftrightarrow$ LLM & Survey and roadmap focused on evolutionary computation in the era of LLMs. & Strongly relevant, but primarily EA-centered and less focused on a broader computer-science synthesis of interaction mechanisms, evaluation issues, and cross-domain design patterns. \\
2025 & Wang et al.~\cite{wang2025large} & LLM $\leftrightarrow$ EA & Discusses conceptual parallels, enhancement opportunities, and challenges when LLMs meet evolutionary algorithms. & Provides a broad perspective, but limited methodological classification and limited critical synthesis of evaluation practice. \\
2026 & Liu et al.~\cite{liu2026systematic} & LLM $\rightarrow$ Algorithm Design & Systematic survey of LLMs for algorithm design, including roles, prompting, search, and applications. & Broader than EC; does not specifically center the bidirectional relationship between EC and LLMs. \\
2026 & Zheng et al.~\cite{zheng2026survey} & LLM $\rightarrow$ Meta-Optimization & Reviews LLM-driven meta-optimizers for automated intelligent optimization. & Focuses primarily on LLMs as meta-optimizers, rather than on the full reciprocal and co-adaptive LLM-EC landscape. \\
\bottomrule
\end{tabular}}
\end{table}

{ \color{black}Motivated by these observations, the novelty of the present survey lies not in chronological priority, but in the specific combination of scope, taxonomy, and methodological emphasis. Relative to the reviews summarized in Table~\ref{tab:comparison}, the present paper differs in four concrete ways. First, it treats the interaction as explicitly bidirectional by covering both EC for LLM enhancement and LLMs for EC enhancement within the same framework, rather than emphasizing only one direction \cite{yu2024deep,wu2025evolutionary,wang2025large,liu2026systematic,zheng2026survey}. Second, it organizes the literature through an EC-aware taxonomy that separates interaction modes by optimization target, functional role, and degree of autonomy, while also clarifying overlap cases and distinguishing between co-adaptive and co-evolutionary systems. Third, it places stronger emphasis on the quality and limits of current empirical evidence, especially with respect to benchmarking, reproducibility, computational cost, contamination, and generalization. Fourth, it integrates emerging co-adaptive systems and algorithm-discovery workflows into a single critical synthesis, thereby positioning LLM-EC interaction as a broader computer-science problem of automated algorithm design rather than merely a collection of isolated methods or applications.

Accordingly, the added value of the present review lies less in covering any single subtopic in isolation than in combining these dimensions into one structured and evaluative synthesis. \color{black}So, the main novelty of the survey lies not in introducing a new, isolated subfield, but in integrating these perspectives into a single EC-aware and methodologically critical account of the current LLM-EC landscape. \color{black}In this sense, the taxonomy proposed in the paper should be read as a structured classification of current interaction patterns rather than as a final or universally fixed ontology of the LLM-EC field.}

The review is intended for two overlapping audiences: (a) researchers in evolutionary computation who seek to leverage LLMs for heuristic generation, surrogate reasoning, algorithm design, or adaptive control; and (b) researchers in AI, optimization, and automated programming who seek to incorporate evolutionary search principles into LLM adaptation, prompting, architecture design, and self-improving systems.

\subsection{Review Scope and Methodology}\label{subsec:review_methodology}
This article is positioned as a \emph{structured narrative review} rather than a fully systematic meta-analysis. The goal is to provide a critical and conceptually organized synthesis of the emerging interaction between LLMs and EC, with emphasis on bidirectional enhancement mechanisms, methodological trends, and open research challenges.

To support this goal, the literature was collected and organized through a retrospective multi-source search procedure. The primary sources included Google Scholar, IEEE Xplore, the ACM Digital Library, SpringerLink, ScienceDirect, and arXiv. Because the area evolves rapidly and several influential works first appeared as preprints, workshop papers, or conference companion papers, the search was not restricted to journal articles alone. Instead, the emphasis was placed on identifying papers that introduced representative frameworks, methodological advances, benchmark-oriented studies, or survey-level discussions relevant to LLM-EC interaction.

The search process combined broad and targeted keyword queries. Representative search terms included combinations of: ``large language model'', ``LLM'', ``GPT'', ``evolutionary computation'', ``evolutionary algorithm'', ``genetic algorithm'', ``differential evolution'', ``genetic programming'', ``swarm intelligence'', ``metaheuristic'', ``hyper-heuristic'', ``prompt optimization'', ``neural architecture search'', ``algorithm design'', and ``meta-optimizer''. Additional backward and forward screening was performed by inspecting the reference lists and citation links of influential papers already identified in the initial search.

 \textcolor{black}{The search primarily focused on publications from 2021–2026, reflecting the rapid emergence of LLM-driven optimization research, while seminal references on evolutionary computation and automated algorithm design were also included to provide historical context. Duplicate records were removed, and the remaining papers were screened based on relevance to the interaction between LLMs and evolutionary computation. Studies that did not contain a substantial methodological contribution, experimental analysis, or conceptual discussion related to the integration of LLMs and EC were excluded. Following the screening process, 72 publications were carefully selected for detailed review.}

The inclusion criteria were as follows: (i) the work addressed LLMs or closely related generative language models in an optimization, search, algorithm-design, or heuristic-design setting; (ii) the work involved EC, evolutionary algorithms, genetic programming, swarm-based search, metaheuristics, or closely related population-based optimization methods; or (iii) the work provided survey, benchmarking, or conceptual material directly relevant to understanding the interaction between these areas. Papers were excluded when they focused only on general NLP applications of LLMs without an optimization component, only on classical EC without relevance to LLMs or automated algorithm design, or on prompt engineering topics that were not meaningfully connected to the LLM-EC interface.

{\color{black}For clarity, the reviewed literature should be interpreted in three broad groups. \emph{Core studies} are those in which the interaction between LLMs and EC is direct and central to the method being proposed or evaluated. \emph{Adjacent studies} are those included because they illustrate a methodological interaction pattern that is directly relevant to the survey, even when the immediate target application is not itself an LLM-centered system. \emph{Historical/background studies} are included selectively to contextualize emerging LLM-EC developments through earlier work on evolutionary search, genetic programming, benchmarking, or automated heuristic generation.

Under this principle, adjacent studies were retained only when they clarified an interaction mechanism that is relevant to the survey taxonomy, whereas background studies were used mainly to support conceptual context rather than to serve as central evidence for current LLM-EC claims.}

Given the recency of the field, both archival and non-archival sources were considered. When a journal or conference version was available, it was preferred over earlier preprints. Preprints were retained when they described influential and actively discussed frameworks for which no archival version was yet available, or when they represented very recent developments that materially affected the positioning of the survey. However, preprint results were interpreted more cautiously than mature archival studies, especially in discussions of empirical evidence, benchmarking, and generalization. {\color{black}Accordingly, areas of the literature that remain dominated by recent or non-archival studies are interpreted in this review as methodologically promising but still provisional, especially where reproducibility, benchmark breadth, or cross-domain validation remain limited.}

The final taxonomy was constructed iteratively after screening the collected literature. Papers were grouped according to the \emph{direction of interaction} between the two paradigms, the \emph{optimization target}, and the \emph{role played by the LLM or EC component} within the overall system. This process resulted in the main categories used throughout the paper: EC for LLM enhancement (including prompt optimization, hyperparameter tuning, and architecture search), LLMs for EC improvement (including surrogate reasoning, operator adaptation, hyper-heuristics, and metaheuristic generation), and emerging co-adaptive LLM-EC systems. This taxonomy was refined further by considering representation level, degree of automation, and methodological maturity.

The purpose of this methodology is therefore not to claim exhaustive coverage of every related publication, but to provide a transparent and reproducible basis for a critical survey of the most relevant and representative developments in this rapidly evolving area. Given the rapid pace of developments in this area, the present review prioritizes representative, methodologically informative, and conceptually influential works rather than claiming exhaustive coverage of every related publication. 

\subsection{Main Contributions}
The main contributions and distinguishing features of this survey are summarized as follows:
\begin{enumerate}[(i)]
\item {\color{black}\textit{Bidirectional and EC-aware synthesis:} We provide a structured review of both directions of interaction, namely EC for enhancing LLMs and LLMs for enhancing EC, while also discussing emerging co-adaptive frameworks and the more specific conditions under which these may be regarded as co-evolutionary.}

\item {\color{black} Structured classification of interaction mechanisms: We organize the literature through an EC-aware classification of current interaction patterns, including prompt and hyperparameter optimization, neural architecture search, surrogate modeling, operator adaptation, hyper-heuristics, metaheuristic generation, and co-adaptive/co-evolutionary pipelines. This framework is intended to clarify how LLMs and EC interact at different representation and control levels, while acknowledging that some category boundaries remain overlapping or historically fluid.}

\item \textit{Critical perspective on evidence and evaluation:} Beyond cataloging methods, we emphasize empirical and methodological issues that are central to a review article, including benchmark selection, reproducibility, generalization, computational overhead, and the difference between proof-of-concept success and robust cross-domain performance.

\item {\color{black}\textit{Cross-cutting interaction contexts and representative applications:} We discuss how LLM-EC synergy appears across optimization, algorithm design, scientific problem solving, and decision-support settings, highlighting representative application contexts and recurring interaction roles.
}

\item \textit{Open problems and research agenda:} We identify major challenges and future directions, including scalable hybrid search, interpretable algorithm generation, safer autonomous design loops, principled benchmarking, and the development of more reliable co-adaptive LLM-EC systems.
\end{enumerate}

{\color{black}Taken together, these contributions and distinguishing features position the survey as a conceptual bridge between evolutionary search and modern generative intelligence. The goal is not merely to summarize an emerging literature, but to provide a clearer foundation for organizing, evaluating, and advancing future research on hybrid LLM-EC systems, with emphasis on interaction mechanisms, methodological critique, and open research challenges.}

The remainder of this paper is organized as follows. Section~\ref{sec: EAs and LLM} examines how EC can be used to enhance LLMs, including prompt optimization, hyperparameter tuning, and architecture search. Section~\ref{sec: LLM and EAs} reviews how LLMs can improve EC through heuristic generation, surrogate reasoning, adaptive operator control, and automated metaheuristic design. Section~\ref{sec: Future scopes} discusses co-adaptive frameworks, open challenges, and future research directions. Finally, Section~\ref{sec: conclusion} concludes the survey.

\section{EC for LLM Enhancement}\label{sec: EAs and LLM}

This section reviews how EC can enhance LLMs when the optimization target lies outside the core pretrained parameters, such as prompts, hyperparameters, architectural configurations, or interaction strategies. In this survey, LLMs are treated primarily as fixed or partially fixed models accessed in a black-box or limited-access manner, which makes gradient-free optimization particularly relevant. From a computer-science perspective, this direction is important because it shows how EC can support the automated design and adaptation of LLM-based systems without relying exclusively on manual tuning or end-to-end retraining. {\color{black}Fig.~\ref{fig:llm_ec_taxonomy_map} positions the major interaction modes surveyed in this paper according to their level of automation and three broad interaction regions: EC for LLM enhancement, LLMs for EC enhancement, and co-adaptive systems. This direction is especially relevant when LLM-based systems expose semantically rich design variables, such as prompts or interaction policies, but must still be optimized under black-box, expensive, or weakly structured evaluation conditions.}

\begin{figure}[h!]
\centering
\begin{tikzpicture}[
    >=Stealth,
    axis/.style={thick, -{Stealth[length=3mm]}},
    cat/.style={draw, rounded corners, align=center, font=\small, inner sep=4pt, fill=white},
    note/.style={font=\small}
]


\draw[dashed, gray] (0,0) -- (14.3,0);
\fill[orange!12] (0,1.4) rectangle (14.2,4.2);
\fill[green!12]  (0,0.05) rectangle (14.2,1.5);
\fill[blue!12]   (0,-3.5) rectangle (14.2,-0.05);

\node[note] at (1.6,4.4) {\textbf{EC for LLM enhancement}};
\node[note] at (1,0.5) {\textbf{Co-adaptive systems}};
\node[note] at (1.6,-3.6) {\textbf{LLMs for EC enhancement}};

\draw[axis] (0,0) -- (15.5,0);
\node[note] at (13.7,0.2) {\small Level of automation};

\node[cat, text width=2.8cm,fill=orange!20] (prompt) at (2.3,3.5) {Prompt\\optimization};
\node[cat, text width=3.0cm,fill=orange!30] (hpo) at (4.8,2.0) {Hyperparameter\\tuning};
\node[cat, text width=3cm,fill=orange!40] (nas) at (7.4,3.2) {Neural architecture\\search (NAS)};

\node[cat, text width=2.8cm,fill=blue!15] (surrogate) at (3.0,-1.5) {Surrogate\\reasoning};
\node[cat, text width=2.8cm,fill=blue!25] (operator) at (5.2,-2.8) {Operator\\control};
\node[cat, text width=3.0cm,fill=blue!35] (hyperheur) at (7.8,-1.5) {Hyper-heuristics/\\program synthesis};
\node[cat, text width=3.1cm,fill=blue!45] (meta) at (10.2,-2.8) {Metaheuristic\\generation};

\node[cat, text width=3.3cm, fill=green!25] (coadaptive) at (10.3,0.9) {Co-adaptive\\LLM-EC systems};

\node[note] at (1.5,-0.45) {\bf low};
\node[note] at (11.3,-0.45) {\bf high};

\draw[dotted, gray] (2.2,0) -- (2.2,3.0);
\draw[dotted, gray] (4.8,0) -- (4.8,1.4);
\draw[dotted, gray] (7.4,0) -- (7.4,2.6);
\draw[dotted, gray] (3.0,0) -- (3.0,-1);
\draw[dotted, gray] (5.2,0) -- (5.2,-2.2);
\draw[dotted, gray] (7.8,0) -- (7.8,-1);
\draw[dotted, gray] (10.2,0) -- (10.2,-2.2);
\draw[dotted, gray] (10.3,0) -- (10.3,0.3);

\end{tikzpicture}
\caption{\color{black}Taxonomy-positioning map of the surveyed interaction modes between LLMs and EC. The horizontal axis indicates increasing levels of automation and autonomy, while the vertical structure is organized into three broad regions: EC for LLM enhancement, LLMs for EC enhancement, and co-adaptive systems as a more integrated interaction zone.}
\label{fig:llm_ec_taxonomy_map}
\end{figure}

Among the different mechanisms available, prompt optimization remains the most visible and mature interface through which EC enhances LLMs. More recently, EC has also been explored for hyperparameter tuning, architecture search, and related design problems in which the search space is large, heterogeneous, or only weakly structured. {\color{black}The following subsections review these developments with emphasis not only on representative methods and search spaces, but also on the empirical maturity of the literature, the strength of available evidence, computational cost, reproducibility, and the extent to which current results support generalizable conclusions.}

\subsection{Prompt Optimization as the Main Interface}

Prompt engineering concerns the design of textual or embedding-based inputs that steer an LLM toward accurate, coherent, and task-appropriate responses. Because LLM outputs can be highly sensitive to prompt wording, structure, and example selection, prompt design has become a central component of downstream adaptation \cite{chen2025unleashing,sahoo2024systematic,clariso2023model}. Manual prompt design, however, is difficult to scale: it is labor-intensive, sensitive to human bias, and often fails to generalize consistently across tasks and datasets \cite{sahoo2024systematic,memon2024llm}. These limitations make prompt optimization a natural target for evolutionary search.

From an optimization perspective, the literature is divided into two broad branches: \emph{hard prompting}, where the search acts on discrete human-readable text, and \emph{soft prompting}, where the optimization acts on continuous prompt embeddings. Hard prompting is currently the more developed branch because it is easier to interpret and more compatible with mutation and crossover operators. Soft prompting remains less explored from an evolutionary perspective and should be viewed as a promising but still immature direction. A concise background note on hard and soft prompting, included only for completeness, is provided in \ref{secapp:appB}.

\subsection{EC in Hard Prompting and Representative Frameworks}\label{subsec:hard_prompting}

Hard-prompt optimization uses EC to search over discrete textual instructions. Early studies such as Genetic Prompt Search (GPS) \cite{xu2022gps} and GrIPS \cite{prasad2023grips} showed that relatively simple gradient-free edits to prompt text can improve task performance. Hsieh et al. \cite{hsieh2023automatic} extended this line by handling longer prompts and preserving useful context across generations. Taken together, these studies established that evolutionary search can replace ad hoc manual prompt crafting with a more systematic optimization process.

A more consequential shift occurred when LLMs themselves became part of the evolutionary machinery. EvoPrompt \cite{guo2024connectinglargelanguagemodels} uses language-model-generated crossover and mutation to create new prompts, while selection remains governed by EC principles. PromptBreeder \cite{fernando2023promptbreeder} further extends this idea by co-evolving both prompts and the mutation instructions used to transform them. Related frameworks, including EvoPrompt \cite{chen2023evoprompting}, PhaseEvo \cite{cui2024phaseevounifiedincontextprompt}, PROMST \cite{chen2024prompt}, EMO-Prompts \cite{baumann2024evolutionary}, GenAP \cite{feng2024genetic}, and PEDO \cite{wong2024generative}, broaden the design space toward multi-phase optimization, human-in-the-loop refinement, multi-objective search, and domain-specific prompt design.

{ \color{black}Among the representative systems discussed above, EvoPrompt, PhaseEvo, and GAAPO were selected for detailed comparison because they illustrate three distinct methodological design strategies in evolutionary prompt optimization: direct LLM-based evolutionary operators, structured multi-phase optimization, and hybrid orchestration of multiple prompt-generation strategies. Table~\ref{tab:comparison2} is therefore intended as a representative comparison rather than an exhaustive inventory of all prompt-optimization frameworks. Other relevant approaches, including PromptBreeder, PROMST, EMO-Prompts, GenAP, and PEDO, remain important and are discussed in the surrounding text as complementary developments. 
Their strengths, limitations, and typical application domains are also included in Table~\ref{tab:comparison2}.}

\begin{table}[h!]
\centering
\caption{\color{black}Comparison of three representative evolutionary prompt optimization frameworks selected to illustrate distinct methodological design strategies: direct LLM-based evolutionary operators (EvoPrompt), structured multi-phase optimization (PhaseEvo), and hybrid orchestration of multiple prompt-generation strategies (GAAPO). The table combines a feature-/mechanism-level comparison with a practical appraisal of strengths, limitations, and application domains.}
\label{tab:comparison2}
\resizebox{1\linewidth}{!}{\begin{tabular}{p{4cm}p{5cm}p{6cm}p{7cm}}
\toprule
 \textbf{Feature} & \textbf{EvoPrompt} & \textbf{PhaseEvo} & \textbf{GAAPO} \\
\midrule
\multicolumn{4}{l}{\textit{Design and mechanism}}\\
\midrule
Primary Goal & Discrete prompt optimization & Unified instruction \& example optimization & Hybrid prompt optimization via strategy integration \\
Core Algorithm & GA/DE & Custom multi-phase EA & GA as controller \\
LLM Role & Implements generic EA operators (crossover, mutation) & Implements phased operators (feedback, EDA, crossover, semantic mutation) & Component within diverse generation strategies (OPRO-like, APO-like, etc.) \\
Optimization Target & Primarily instruction & Joint instruction \& examples & Primarily instruction (with few-shot generator for examples) \\
Key Operators/Strategies & LLM-based crossover, LLM-based mutation & Feedback mutation, EDA mutation, crossover mutation, semantic mutation & OPRO-like, APO-like, random mutator, crossover, few-shot \\
Evaluation & Development-set score & Development-set score & Flexible: complete, successive halving, bandit selection \\
Notable Innovations & LLM as direct EA operator; gradient-free black-box optimization & Unified ICL optimization; multi-phase structure; task-aware similarity metric & Hybrid integration of multiple APO strategies; flexible evaluation; trade-off analysis \\
\midrule
\multicolumn{4}{l}{\color{black}\textit{Practical appraisal}}\\
\midrule
{\color{black}Strengths} & {\color{black}Interpretable, gradient-free, black-box compatible, and suitable for discrete prompt optimization} & {\color{black}Unified, multi-phase, task-aware, and well-suited to instruction--example optimization} & {\color{black}Integrates diverse optimization strategies, supports flexible evaluation, and can improve robustness across tasks} \\
{\color{black}Limitations} & {\color{black}Performance may vary across tasks, depends strongly on LLM capability, and may suffer from prompt drift} & {\color{black}Implementation complexity, high API cost, and sensitivity to initialization} & {\color{black}Computationally expensive; may overfit with large populations; depends on the quality of integrated strategies} \\
{\color{black}Application Domain} & {\color{black}Sentiment and topic classification, summarization, and reasoning tasks (e.g., BBH)} & {\color{black}BBH reasoning, instruction induction, and sarcasm detection} & {\color{black}Hate speech detection, professional benchmarks, and graduate-level Q\&A} \\
\bottomrule
\end{tabular}}
\begin{flushleft}\color{black}\footnotesize \textit{Note:} EA: Evolutionary algorithm; GA: Genetic algorithm; DE: Differential evolution; ICL: In-context learning; APO: Automatic prompt optimization; OPRO: Optimization by prompting; EDA: Estimation-of-distribution algorithm; BBH: Big-Bench Hard. Strengths, limitations, and application domains (previously the separate appendix table) have been merged into this table.
\end{flushleft}
\end{table}

Overall, the literature on hard prompting shows a clear progression: from token-level search over discrete prompts to hybrid frameworks in which LLMs actively generate, mutate, and refine candidate prompts inside an evolutionary loop. This progression is important to the survey's broader theme because it illustrates a concrete form of LLM-EC synergy: EC provides structured search and selection pressure, while the LLM provides semantic flexibility and linguistic variation.

{\color{black}From a critical perspective, hard-prompt evolution currently has a stronger empirical base than most other EC-for-LLM directions. However, the available evidence remains concentrated in relatively narrow task families and benchmark settings. Many reported gains are promising, yet cross-task robustness, sensitivity to evaluation design, and reproducibility across models and prompt initializations remain insufficiently documented.}

\subsection{EC in Soft Prompting}\label{subsec:soft_prompting}

Soft prompting, also known as prompt tuning, optimizes continuous prompt embeddings rather than explicit text. In contrast to hard prompts, these representations operate directly in the embedding space of the model and can provide compact task adaptation. Most existing soft-prompt methods rely on gradient-based optimization, reinforcement learning, or sequential optimal learning rather than EC \cite{zhu2023prompt,zhang2022tempera,wang2025sequential}.

The limited use of EC in soft prompting is not accidental. Evolutionary search in high-dimensional embedding spaces poses several difficulties, including designing meaningful mutation and crossover operators, avoiding degenerate or adversarial embeddings, incurring the cost of repeated model inference, and defining fitness measures that capture subtle performance differences. These issues make soft-prompt evolution considerably harder to interpret and often more expensive than text-level prompt search.

{\color{black}
We emphasize that these observations concern the current maturity and interpretability of the evolutionary soft-prompting literature rather than an inherent limitation of EC in continuous spaces: population-based search is, in principle, well-suited to non-convex and multimodal embedding spaces, and the present gap is primarily empirical, which reflects underdeveloped operators and limited comparative validation.
}

Despite these limitations, soft prompting remains a promising area for EC. Population-based search is naturally suited to non-convex and multimodal spaces, and hybrid strategies that combine EC with gradient-based updates may improve exploration while controlling overfitting. Surrogate-assisted search, low-dimensional encodings, and co-evolutionary schemes that jointly optimize hard and soft prompts also appear promising. For this survey, soft prompting should be viewed less as a mature EC application area than as an open methodological frontier for future hybrid search strategies.

{\color{black}At present, however, the literature in this area is better interpreted as a methodological opportunity than as an empirically validated direction, since reproducible evidence against strong gradient-based or hybrid baselines remains very limited.}


\subsection{Evolutionary Hyperparameter Tuning for LLMs}\label{subsec:hyperparameter}
Beyond prompt optimization, EC has also been used to tune a broader set of design variables in LLM-centered pipelines, including architectural hyperparameters, control parameters, and model-composition settings. This broader view of hyperparameter tuning is especially relevant when LLMs are accessed as black-box systems or when the design space contains interacting variables that are expensive to optimize manually.

The main appeal of EC in this setting is its gradient-free, population-based search. Rather than optimizing a single configuration locally, EC can explore multiple candidate settings in parallel and capture interactions among hyperparameters that simpler search methods may miss. Table~\ref{tab:ea_llm_studies} summarizes representative studies, while a reorganized synthesis by hyperparameter category is moved to \ref{app:hpo_supplement}.

At the same time, the literature remains relatively scattered. Some studies focus on architecture-related hyperparameters, such as layer counts and hidden dimensions, whereas others target model merging, self-adaptation, or more general machine-learning hyperparameters. The main takeaway is not the number of available examples, but the recurring pattern: EC is most useful when the search space is difficult to model explicitly, gradients are unavailable or expensive, and multiple competing objectives, such as performance, efficiency, and latency, must be balanced. Compared with prompt optimization, however, the empirical literature on EC-based hyperparameter tuning for LLM-centered systems remains limited, and the available evidence is still fragmented across heterogeneous optimization settings.

{\color{black}{
To clarify when EC is preferable, it is useful to position it against the main alternatives for hyperparameter and configuration search in LLM-centered pipelines. Random search is a simple, naturally parallel baseline that is often surprisingly competitive, but it is uninformed and exploits neither evaluation history nor interactions among variables. Grid search is exhaustive and interpretable, yet scales poorly with dimensionality and wastes evaluations on irrelevant regions. Bayesian optimization is highly sample-efficient for low-dimensional, continuous, and expensive objectives, but it is largely sequential and becomes difficult to apply in high-dimensional, conditional, or mixed discrete-continuous spaces \cite{hao2024model, aglietti2024funbo}. Reinforcement learning can learn adaptive search policies, but is typically sample-inefficient, sensitive to reward design, and costly to stabilize. Gradient-based optimization is efficient when gradients or differentiable relaxations are available, but is inapplicable to black-box LLM access, discrete architectural choices, or non-differentiable objectives \cite{zhang2022tempera}. Against this backdrop, EC occupies a distinct niche: it is gradient-free, population-based, naturally parallel, and well-suited to heterogeneous, mixed, multimodal, and multi-objective spaces with interacting variables. Its principal cost is the large number of evaluations required, which is precisely why EC is frequently combined with surrogate-assisted search rather than treated as an alternative to it \cite{hao2024large}. In short, EC is most attractive when the configuration space is black-box, heterogeneous, and multi-objective, whereas Bayesian optimization, gradient-based methods, or simple random search may dominate in lower-dimensional, differentiable, or cheaply evaluable settings. More broadly, the relevance of EC to LLM-centered hyperparameter and configuration search is consistent with the wider literature on hierarchical and bilevel optimization, where evolutionary algorithms and learning-based methods have been studied as complementary tools for tackling nested, black-box design problems \cite{11518551}.

}}

{ \color{black}Critically, this area remains less mature than prompt optimization. The available studies suggest that EC can be useful when the design space is heterogeneous, black-box, or multi-objective, but the evidence base is still too fragmented to support broad claims of superiority over alternative tuning strategies across LLM-centered settings.}

\begin{table}[h!]
\caption{Representative studies on EA-based hyperparameter optimization for LLM-centered pipelines.}
\label{tab:ea_llm_studies}
\centering
\resizebox{1\linewidth}{!}{\begin{tabular}{p{0.20\textwidth}p{0.12\textwidth}p{5cm}p{0.15\textwidth}p{6cm}}
\toprule
\textbf{Study/Method} & \textbf{EA used} & \textbf{Targeted Hyperparameters} & \textbf{LLM/Task Context} & \textbf{Key Finding/Contribution}\\
\midrule
AutoTinyBERT~\cite{yin2021autotinybertautomatichyperparameteroptimization} & Custom EA (Evolver) & Architectural dimensions, layers ($l_t$, $d_m$, etc.) & BERT/efficiency (latency) & Automated architectural HPO using EA and a proxy model for efficient PLMs. \\
Evolutionary Merging~\cite{akiba2025evolutionary} & CMA-ES & Model-merging recipe (TIES/DARE) & Foundation model merging & EA automates the discovery of effective merging hyperparameters. \\
Tani et al.~\cite{tani2021evolutionary} & GA, PSO & General ML hyperparameters & ML for high-energy physics & Explored GA/PSO for autonomous HPO in a domain-specific setting. \\
\bottomrule
\end{tabular}}
\begin{flushleft}
\color{black}\footnotesize \textit{Note:} EA: Evolutionary algorithm; CMA-ES: Covariance matrix adaptation evolution strategy; GA: Genetic algorithm; PSO: Particle swarm optimization; HPO: Hyperparameter optimization; PLM: Pretrained language model.
\end{flushleft}
\end{table}

\subsection{Evolutionary Architecture Optimization for LLMs}\label{subsec:NAS}

Neural architecture search (NAS) provides another important avenue for EC to enhance LLMs. The appeal of EC in this setting lies in its ability to handle heterogeneous, partly discrete design spaces while balancing multiple objectives, including accuracy, latency, memory, and parameter efficiency. These properties make evolutionary NAS particularly relevant for compressed, specialized, or resource-constrained LLM variants.

Evolutionary NAS has been applied at different abstraction levels, including overall backbone discovery, macro-level structural hyperparameters, layer configuration, pruning, and code-level modifications. {\color{black}Some studies in this category, such as LLM-guided code-level design for graph neural networks, are included here as \emph{adjacent} rather than core LLM-architecture studies. Their relevance lies not in optimizing the internal architecture of an LLM itself, but in illustrating a broader methodological pattern in which language-guided evolutionary search is used to generate or refine candidate architectures at the code level. Accordingly, this subsection includes both core LLM-related NAS studies and a small number of adjacent code-level studies that help clarify this broader interaction pattern.} In some cases, the LLM participates directly by generating or mutating code-level architectural variants; in others, the architecture is searched through more conventional EC encodings. {\color{black}Table~\ref{tab:nas_methods} summarizes the main categories of evolutionary NAS relevant to LLM-related settings, while a more detailed study-wise summary is provided in~\ref{app:nas_supplement}.}

From a review perspective, the main point is that EC is valuable here not simply because it searches broadly, but because it can accommodate heterogeneous design choices and multiple objectives more naturally than many gradient-based approaches. However, evolutionary NAS for LLMs still faces a severe evaluation bottleneck, since even an approximate assessment of candidate architectures can be expensive. This limitation remains a central barrier to wider adoption.

{\color{black}The critical limitation of evolutionary NAS in this context is therefore not only computational cost, but also evidentiary depth. Many studies demonstrate methodological feasibility, but far fewer establish robust comparative advantage under realistic evaluation budgets, model scales, or deployment constraints.}

\begin{table}[h!]
\centering
\caption{Overview of evolutionary NAS directions in LLM-related settings.}
\label{tab:nas_methods}
\resizebox{1\linewidth}{!}{%
\begin{tabular}{>{\raggedright\arraybackslash}p{4cm}>{\raggedright\arraybackslash}p{12cm}>{\raggedright\arraybackslash}p{5cm}}
\toprule
\textbf{Category} & \textbf{Description} & \textbf{Representative Examples} \\
\midrule
\multicolumn{3}{l}{\textbf{Optimized Architectural Aspects}} \\
\midrule
Overall Structure/Backbone &
Discovery of fundamental model architectures, including backbone design and high-level attention structure &
AutoBERT-Zero~\cite{gao2022autobert} \\

Macro-Level Hyperparameters &
Optimization of high-level structural choices such as the number of blocks, hidden dimensions, attention heads, and FFN sizes &
SuperShaper~\cite{ganesan2021supershaper}, AutoTinyBERT~\cite{yin2021autotinybertautomatichyperparameteroptimization} \\
Component Choices & \textcolor{black}{Search over discrete module-level design choices within model components; (e.g., operation/primitive selection such as attention vs. convolution, and activation functions) }
 & Evolved Transformer \cite{so2019evolved}.
 \\
Layer Configuration &
Optimization of layer-specific hyperparameters and connectivity patterns &
LiteTransformerSearch~\cite{javaheripi2022litetransformersearchtrainingfreeneuralarchitecture} \\

Code-Level/LLM-Assisted Modifications &
LLM-guided or code-level generation and refinement of candidate architectures in evolutionary search pipelines &
LLMatic~\cite{nasir2024llmatic}, EvoPrompting~\cite{chen2023evoprompting} \\
\midrule
\multicolumn{3}{l}{\textbf{Evolutionary NAS Methods and Techniques}} \\
\midrule
Standard EC (GA/NSGA-II) &
Classical evolutionary search, including single-objective and multi-objective variants &
AutoBERT-Zero~\cite{gao2022autobert}, DistilBERT NAS~\cite{paraskeva2024resource} \\

Multi-Objective EAs (MOEAs) &
Joint optimization of competing objectives such as accuracy, latency, memory, and model size &
LiteTransformerSearch~\cite{javaheripi2022litetransformersearchtrainingfreeneuralarchitecture} \\

QD &
Search for diverse high-performing solutions rather than a single optimum &
LLMatic~\cite{nasir2024llmatic} \\

LLM-Guided Evolution &
Use of LLMs as variation or generation operators within an evolutionary architecture-search loop &
EvoPrompting~\cite{chen2023evoprompting} \\
\bottomrule
\end{tabular}}
\begin{flushleft}
\color{black}\footnotesize \textit{Note:} NAS: Neural architecture search; GA: Genetic algorithm; NSGA-II: Non-dominated sorting GA II; MOEA: Multi-objective evolutionary algorithm; QD: Quality-diversity.
\end{flushleft}
\end{table}

\subsection{Current Challenges and Future Directions}
{\color{black}Taken together, the literature reviewed in this section suggests an uneven maturity profile across EC-for-LLM directions. Prompt optimization currently has the strongest and most visible evidence base, particularly in hard-prompt settings where the search space is discrete, interpretable, and naturally compatible with evolutionary variation operators. By contrast, EC-based hyperparameter tuning remains more fragmented across heterogeneous optimization settings, and evolutionary NAS for LLM-related systems is still constrained by severe evaluation cost and limited large-scale comparative evidence.}

Despite this promise, the field remains constrained by several recurring bottlenecks. The first is evaluation cost: prompt candidates, hyperparameter settings, and architectural variants often require repeated model inference, partial retraining, or proxy evaluation, which makes large-scale evolutionary search expensive. {\color{black}This issue is especially acute in architecture search, where even approximate assessment is costly, as in
AutoTinyBERT~\cite{yin2021autotinybertautomatichyperparameteroptimization}
and the training-free workaround adopted by LiteTransformerSearch~\cite{ javaheripi2022litetransformersearchtrainingfreeneuralarchitecture}; prompt-level search such as EvoPrompt~\cite{ guo2024connectinglargelanguagemodels} similarly incurs repeated inference per candidate.} A second challenge is representation. Prompts, embeddings, and architectures require different encodings, and poorly aligned representations can distort search behavior or limit the usefulness of mutation and crossover. {\color{black}This is particularly visible in soft-prompt
evolution (Section~\ref{subsec:soft_prompting}), where operator design over
continuous embeddings remains underdeveloped \cite{bala2025orpsoc}, and in large architecture spaces
such as AutoBERT-Zero~\cite{ gao2022autobert} and
LLMatic~\cite{nasir2024llmatic}, where encoding choices strongly shape search
behavior.} A third challenge concerns reliability and transferability, since approximate or surrogate fitness estimates may not generalize consistently across tasks, datasets, model scales, or evaluation setups \cite{chang2024survey}. {\color{black}For example, the proxy- and surrogate-based
fitness used to contain cost (as in training-free NAS~\cite{javaheripi2022litetransformersearchtrainingfreeneuralarchitecture}) may not transfer consistently across these conditions.}

For these reasons, future work should focus less on simply expanding the number of hybrid methods and more on improving their methodological foundations. Particularly important directions include: (i) low-cost and reliable fitness evaluation, including surrogate-assisted and multi-fidelity search; (ii) better encodings and operators for continuous prompts and large architectures; (iii) hybrid EC--gradient or EC--LLM methods that combine broad exploration with local refinement; (iv) stronger evaluation protocols addressing robustness, reproducibility, and generalization across tasks and models; and (v) more principled treatment of practical constraints such as latency, safety, and resource usage.

In summary, EC offers a compelling toolbox for improving LLM systems when the optimization problem is black-box, discrete, multimodal, or multi-objective. Its strongest current impact is in prompt optimization, while hyperparameter tuning and evolutionary NAS remain important but methodologically less mature directions. As the literature develops, the key challenge will be to move from proof-of-concept demonstrations toward scalable, reproducible, and better-validated EC-based adaptation pipelines for LLMs.
\section{LLMs for EC Enhancement}\label{sec: LLM and EAs}
This section reviews how LLMs enhance EC at progressively higher levels of autonomy. At the most conservative level, LLMs assist existing EC pipelines by approximating fitness, tuning parameters, or extracting useful structural patterns from observed search behavior. At a more advanced level, they act as hyper-heuristic engines that generate, refine, or select search rules. At the highest level, they can propose complete metaheuristics or directly mediate parts of the evolutionary search process itself. This progression is important for understanding the broader significance of LLMs in EC: their role is not limited to text generation, but extends to reasoning, abstraction, program synthesis, and adaptive control. Implementation-oriented supplementary details supporting selected parts of this section are collected in \ref{secapp:appA}.

From a methodological perspective, these interactions can be understood at three levels of representation. Numerical representations correspond to parameter values and continuous controls; symbolic representations encode algorithmic structures such as trees, operators, or programs; and textual representations describe heuristics or search procedures in natural language. Much of the recent literature on LLM-enhanced EC can be interpreted as moving between these levels: LLMs describe or generate algorithmic ideas in text, these ideas are instantiated in symbolic or executable form, and their performance is assessed numerically through search outcomes. This multi-representational perspective helps clarify why LLMs are particularly well-suited to high-level design tasks within EC.

{\color{black}Because several recent frameworks combine multiple roles within a single pipeline, the categories used in this section should be interpreted as a structured organizing framework rather than as fully disjoint classes or a final fixed taxonomy. In cases of overlap, a method is assigned to the category that best reflects the \emph{primary functional role} played by the LLM in the evolutionary process. This classification is based primarily on three criteria: the object being generated or modified (e.g., a heuristic component, surrogate judgment, an operator-control signal, or a full algorithm), the level of autonomy attributed to the LLM, and the dominant methodological contribution emphasized in the original work. Under this principle, overlaps are acknowledged explicitly, but each framework is discussed in the subsection that most clearly captures its main role in the LLM-EC interaction.}

{\color{black}In line with this organizing framework, the discussion in this section remains evaluative rather than merely descriptive, with particular attention to differences in autonomy, reproducibility, empirical support, and the extent to which current frameworks go beyond proof-of-concept demonstrations.}

\subsection{LLM-Guided Heuristic and Metaheuristic Design}\label{subsec: LLM for EC}
An early and influential line of work uses LLMs to propose new search operators or hybrid metaheuristics from textual prompts. Pluháček et al.~\cite{pluhacek2023investigating} showed that GPT-4 can generate new mutation strategies for differential evolution (DE), demonstrating that language models can contribute nontrivial algorithmic variation rather than merely describing existing methods. In related work, {\ \color{black}Pluháček et al. \cite{pluhacek2023leveraging} used GPT-4 to compose hybrid swarm-intelligence algorithms by recombining elements from particle swarm optimization (PSO), cuckoo search (CS), artificial bee colony (ABC), grey wolf optimizer (GWO), the self-organizing migrating algorithm (SOMA), and the whale optimization algorithm (WOA).} A subsequent extension applied repeated prompting to refine the Self-Organizing Migrating Algorithm (SOMA), further suggesting that LLMs can iteratively improve metaheuristic design when performance feedback is available \cite{pluhacek2024using}. 

Taken together, these studies indicate that LLMs can function as meta-level generators of search strategies. Their main contribution lies in semantic recombination: rather than searching directly over numeric solutions, the model searches over descriptions of operators, rules, and algorithmic structures. This capability is attractive for EC because it can reduce the manual effort required to design new metaheuristics, though the resulting methods still require careful empirical validation and benchmarking. 

{\color{black}From a critical perspective, the evidence in this area is currently more convincing for proof-of-concept semantic recombination than for robust methodological advance. Existing studies show that LLMs can propose nontrivial operator variants and hybrid metaheuristic structures, but far fewer establish reproducible gains over carefully tuned classical design workflows or expert-crafted baselines across diverse problem classes. As a result, the present value of this literature lies more in demonstrating a new mode of algorithmic assistance than in confirming broad performance superiority.}

\subsection{LLM-Based Hyper-Heuristic and Program-Synthesis Frameworks}\label{subsubsec: LLM and HHs}
A more structured direction treats LLMs as hyper-heuristic engines. Unlike a conventional metaheuristic, which operates directly in the solution space, a hyper-heuristic operates over the space of heuristics, selecting, generating, or refining search rules. {\color{black}In this review, frameworks in this category are classified primarily by the fact that the LLM operates at the level of heuristic components, search rules, controller logic, or executable heuristic fragments, rather than by producing a complete standalone optimization algorithm.} In this context, LLMs are appealing because they can express search logic in natural language, translate it into code, and revise it based on historical performance.

A landmark example is \texttt{FunSearch} \cite{romera2024mathematical}, which combines LLM-based code generation with iterative evaluation and selection. Candidate code snippets are proposed by the model, scored by a task-specific evaluator, and recycled into future generations. This closed-loop design inspired related systems such as \textit{FunBO} \cite{aglietti2024funbo}, which evolves Bayesian-optimization acquisition functions, and \textit{Evolution of Heuristics (EoH)} \cite{liu2024evolution}, which jointly represents heuristic ideas in natural language and executable code. EoH is particularly notable because it makes the dual role of language explicit: LLMs do not simply emit code, but also articulate the rationale behind candidate heuristics.

Subsequent extensions strengthen this paradigm. \textit{MEoH} \cite{yao2025multi} generalizes EoH to multi-objective settings by maintaining diversity across both solution quality and heuristic form. \texttt{ReEvo} \cite{ye2024reevo} adds reflective feedback, distinguishing between short-term reflections tied to recent parents and long-term reflections accumulated across the search history. \textit{LLaMEA} \cite{van2025llamea} integrates GPT-4 into an EA-style outer loop and evaluates generated algorithms systematically with the \texttt{IOHprofiler} suite \cite{doerr2018iohprofiler,de2024iohexperimenter,wang2022iohanalyzer}, while \textit{LLaMEA-HPO} couples the same idea with Bayesian optimization for hyperparameter refinement \cite{van_Stein_2025}.

These frameworks mark an important shift in EC research: instead of handcrafting heuristics and then tuning them, the search process can increasingly generate and refine the heuristics themselves. For a review article, the main contribution of this literature is conceptual as much as practical. It suggests that LLMs may serve as a new layer of abstraction in EC, one in which algorithm design, explanation, and adaptation can be partially unified inside a single search loop.
At the same time, much of this literature remains recent and, in several cases, heavily dependent on proprietary LLMs, benchmark-specific evaluation, or limited ablation evidence, which makes reproducibility and cross-domain generalization important open concerns. 

{\color{black}To improve reproducibility in this area, future studies should report prompts, model versions, seeds, inference settings, and evaluation budgets more systematically, and should release code and generated heuristics whenever possible. Where feasible, evaluation with open-weight models, stronger ablations, and held-out testing would further help distinguish genuine methodological gains from dependence on a particular proprietary model setup. \color{black}As a result, current evidence is more convincing in showing that LLMs can participate meaningfully in heuristic-generation loops than in establishing that these frameworks consistently outperform well-tuned classical design strategies across problem classes.}

\subsection{LLM-Assisted EC Tuning and Surrogate Reasoning}\label{subsec:LLM-assisted EC}

Not all LLM-based contributions aim to design new heuristics. A second line of work improves existing EC algorithms by using LLMs as surrogates, tuning agents, or pattern extractors. This direction is especially relevant when objective evaluations are expensive, search behavior is difficult to interpret numerically, or adaptation decisions benefit from textual reasoning over historical search data.

\subsubsection{Surrogate Modeling for EA Optimization}

Hao et al.~\cite{hao2024large,hao2024model} proposed LLM-assisted EA (LAEA), where surrogate modeling is reformulated as an inference problem. Instead of fitting a conventional surrogate model, the LLM is prompted with historical candidate-fitness pairs and asked to estimate the quality of new candidates through classification or regression. In parallel, LLAMBO \cite{liu2024large1} integrates LLMs into Bayesian optimization by converting prior configuration-performance information into textual context, thereby helping initialize surrogates and guide sampling under limited data.

The significance of these studies lies less in outperforming all classical surrogates and more in showing that LLMs can act as knowledge-rich approximators when structured numerical history is converted into text. This idea is intriguing for EC because it broadens the notion of surrogate assistance: the surrogate need not be purely statistical, but can also leverage semantic inference from contextualized search trajectories. A generic workflow and an illustrative regression-style prompt are provided in \ref{app:surrogate_workflow} and \ref{app:regression_prompt}, respectively.

\subsubsection{Adaptive Operator and Parameter Control via LLMs}
LLMs have also been used to adapt the behavior of existing metaheuristics online. Martinek et al.~\cite{martinek2024large} studied LLM-driven parameter refinement for algorithms such as GA, ACO, PSO, and SA on problems including the TSP and graph coloring. Here, the LLM receives summaries of the problem, current parameters, and preliminary performance statistics and returns updated parameter suggestions. This effectively turns the model into an adaptive controller that reasons over optimization history rather than relying on fixed update rules.

This direction is especially relevant because it highlights a broader design pattern: LLMs can sit above an existing EC algorithm as an interpretable control layer. {\color{black}The strength of this approach is flexibility; its main limitation is limited reliability, since performance depends heavily on prompt formulation, the fidelity of the summary statistics, and the consistency of the model’s recommendations across runs.} Representative prompt templates used in this style of LLM-guided tuning are listed in \ref{app:tsp_prompts}.

\subsubsection{Pattern-Guided Evolution}

A related but more problem-structured approach is illustrated by \texttt{OptiPattern} \cite{sartori2024metaheuristics}, which augments a biased random key genetic algorithm (BRKGA) with LLM-derived structural priors for the multi-hop influence maximization problem. Rather than directly generating solutions, the LLM infers node-selection probabilities from graph descriptions, and these probabilities bias the decoder toward promising regions of the search space.

This example is important because it shows that LLMs can contribute to EC without replacing the search process itself. Instead, they inject domain-structured bias into the evolutionary loop. More broadly, this suggests a practical middle ground between purely hand-designed heuristics and fully LLM-generated algorithms.

{\color{black}From a critical standpoint, the studies discussed throughout this section are important primarily because they expand the role of LLMs within EC pipelines, not because they have yet displaced classical surrogates or adaptive controllers in a broadly validated way. Their current value is therefore more methodological and exploratory than decisively performance-driven.}

\subsection{LLM-Generated Metaheuristics}\label{subsec: LLM and MHs}
At the highest level of autonomy, LLMs are used not only to guide or tune EC algorithms, but to generate complete metaheuristics. This direction differs from the hyper-heuristic literature in one key respect: the target is no longer a heuristic component, controller, or local search rule, but an entire search procedure with its own inspiration, logic, and update rules. {\color{black}Accordingly, methods are classified here as LLM-generated metaheuristics when the dominant output is a full algorithmic search procedure rather than a partial heuristic element within a broader design loop.}

\subsubsection{Metaheuristic Generation via Structured Prompting}\label{subsec: LLM for metaheuristic}

A representative example is Zoological Search Optimization (ZSO) \cite{zhong2024leveraging}, which was generated by ChatGPT-3.5 under the CRISPE prompting framework. The method illustrates how structured prompting can produce a full algorithmic template, including inspiration, equations, parameter definitions, and flowchart-style logic. Its importance lies less in the specific zoological metaphor and more in what it demonstrates: LLMs can act as autonomous algorithm inventors when given a sufficiently scaffolded design prompt.

\subsubsection{Language-Mediated Evolutionary Search}

Liu et al.~\cite{liu2024large} proposed LMEA, a zero-shot framework in which the LLM itself performs selection, crossover, and mutation through prompt-based reasoning. Rather than treating the LLM as an auxiliary assistant, LMEA embeds it directly inside the evolutionary loop. Candidate solutions and their fitness values are presented as in-context examples, and the model is instructed to generate offspring in a structured format. A self-adaptive temperature mechanism is used to moderate exploration and exploitation over time.

LMEA is conceptually significant because it blurs the line between evolutionary algorithms and language models. Instead of manually specifying evolutionary operators, the search process is described in language and enacted by the model itself. This makes LMEA a useful case study for the future of language-mediated optimization, although its scalability, stability, and reproducibility still require careful study. A detailed example of the LMEA prompt structure is provided in \ref{app:lmea_prompt}.

Taken together, the literature on LLM-generated metaheuristics suggests a continuum of autonomy. At one end, LLMs suggest changes to an existing algorithm; at the other, they instantiate the algorithmic search procedure itself. This continuum is useful for organizing the field because it makes clear that “LLMs for EC” is not a single methodological category but a layered design space. { \color{black}This makes the literature on LLM-generated metaheuristics conceptually significant but empirically unsettled. In many cases, claims of novelty are easier to make than to validate, and the evidence remains stronger for the feasibility of generation than for the robust demonstration of algorithmic originality or broad performance advantage.  \color{black}In particular, future studies in this area should distinguish more clearly between novelty in presentation and in the search mechanism, and should validate the latter through matched-budget comparisons, traceability to known algorithm families, and contamination-aware assessment.}

\subsection{Genetic Programming and Learning-Enhanced Evolution}

The interaction between LLMs and EC also connects naturally to genetic programming (GP) and learnable evolution models, both of which emphasize the generation of structured search rules rather than direct numerical optimization. Although not all work in this area uses modern LLMs explicitly, it provides an important bridge between classical program evolution and emerging language-driven algorithm design.

\subsubsection{GP-Based Generative Hyper-Heuristics}
Zhang et al.~\cite{zhang2021multitask} introduced a multitask GP-based generative hyper-heuristic framework for dynamic scheduling, where the goal is not merely to select among existing heuristics but to evolve new ones across related tasks. Zhang et al.~\cite{zhang2022multitask} extended this idea to a multitask multi-objective setting for dynamic flexible job shop scheduling, using task-aware crossover to preserve task relevance while enabling cross-task knowledge sharing.

These studies are relevant here because they show that heuristic generation and knowledge transfer were already important themes in EC before the arrival of modern LLMs. LLM-based systems inherit and extend this perspective by replacing or augmenting symbolic program evolution with language-guided synthesis and explanation.

\subsubsection{Learnable Evolution Models}
Dashti et al.~\cite{dashti2022lemabe} proposed LEMABE, which combines a learnable evolution model with analogy-based estimation for software cost prediction. The framework alternates between evolutionary search and learning-guided refinement of feature weights, thereby illustrating a broader principle: evolutionary search becomes more efficient when the generation of new candidates is informed by learned patterns rather than solely by blind variation.

For the present survey, the main value of such models is conceptual: they help position modern LLM-enhanced EC within a longer trajectory of research on intelligent variation, learned priors, and automated heuristic design. {\color{black}Critically, this bridge literature is more valuable as conceptual context than as direct empirical evidence for the current generation of LLM-enhanced EC systems. It shows that intelligent variation, heuristic generation, and learnable search bias all have deeper roots in EC, but it does not by itself validate stronger contemporary claims about LLM-driven algorithm design, cross-domain generalization, or consistent superiority over established baselines. Its main contribution to the present review is therefore interpretive: it helps situate recent LLM-based developments within a longer methodological lineage rather than serving as decisive evidence for their effectiveness.}

\subsection{Synthesis and Perspective}
The literature reviewed in this section shows that LLMs improve EC in at least three distinct ways. {\color{black}These categories are analytically useful but not always fully disjoint, since several frameworks combine heuristic generation, reflective revision, surrogate assistance, and algorithm-level design within a single pipeline. For this reason, the preceding classification should be interpreted according to the primary functional role assigned to the LLM in each case. More broadly, it should be read as a structured classification of current interaction patterns rather than as a set of rigid methodological partitions.} First, they provide \emph{reasoning-aware assistance} through surrogate modeling, parameter adaptation, and pattern extraction. Second, they provide \emph{heuristic-level abstraction} through hyper-heuristics and program-synthesis frameworks that generate and refine search rules. Third, they enable \emph{algorithm-level generation}, in which complete metaheuristics or language-mediated evolutionary procedures are generated from prompts and feedback.

{\color{black}In particular, the main distinction across the preceding subsections is between frameworks that generate or refine heuristic elements inside a broader search process and those whose dominant output is a full search procedure. Algorithm discovery is discussed in Section~\ref{sec: Future scopes} as the broader research agenda that spans both cases.}

{\color{black}Taken together, however, the strength of empirical support remains uneven across these categories. The strongest evidence is for assistance-oriented and heuristic-generation roles, where LLMs meaningfully contribute to design, adaptation, or search support without fully replacing the evolutionary process. By contrast, the evidence is less mature for broader claims about robust algorithmic discovery, reproducibility across problem classes, and consistent superiority over strong classical baselines.}

The most important conclusion is that LLM-enhanced EC should not be judged solely by whether a single generated heuristic outperforms a baseline on a benchmark. Its broader significance lies in how it changes the workflow of algorithm design: from manually specified search rules toward systems that can propose, explain, revise, and sometimes implement those rules automatically. At the same time, this shift raises open questions about reliability, evaluation standards, generalization, reproducibility, and the extent to which linguistic reasoning can substitute for principled algorithmic design. Table~\ref{tab: llm_evolution_summary} provides a compact summary of representative frameworks discussed in this section. {\color{black}A broader comparative synthesis across the major interaction modes, including performance evidence, computational burden, reproducibility, and generalization, is provided later in Table~\ref{tab:comparative_synthesis}.}

\begin{table}[ht!]
\centering
\caption{Representative frameworks for LLM-enhanced EC, organized by their primary functional role in the evolutionary pipeline.}
\label{tab: llm_evolution_summary}
\resizebox{1\linewidth}{!}{%
\begin{tabular}{p{3.2cm}p{4.1cm}p{9.5cm}p{5.2cm}}
\hline
\textbf{Framework} & \textbf{Primary Role} & \textbf{Core Idea} & \textbf{Main Relevance to This Survey} \\
\hline
SOMA/SOMA-ATA \cite{pluhacek2024using,pluhacek2023leveraging} 
& LLM-guided metaheuristic refinement 
& Uses LLMs to generate or revise search operators and hybrid swarm-intelligence strategies 
& Illustrates early LLM-assisted algorithm design \\ \hline

EoH/MEoH \cite{liu2024evolution,yao2025multi} 
& Hyper-heuristic generation 
& Evolves heuristics through a combination of natural-language reasoning and executable code 
& Shows how LLMs can search over heuristic space rather than solution space \\ \hline

\texttt{ReEvo} \cite{ye2024reevo} 
& Reflective heuristic evolution 
& Uses short-term and long-term reflection to iteratively improve generated heuristics 
& Introduces meta-cognitive feedback into EC design \\ \hline

ZSO \cite{zhong2024leveraging} 
& Full metaheuristic generation 
& Generates an entire optimization algorithm through structured prompting 
& Demonstrates autonomous algorithm invention from language prompts \\ \hline

LMEA \cite{liu2024large} 
& Language-mediated evolutionary search 
& Uses an LLM to perform selection, crossover, and mutation directly through prompting 
& Blurs the boundary between EA operators and language generation \\ \hline

LLaMEA \cite{van2025llamea} 
& Algorithm generation and refinement 
& Embeds GPT-4 in an EA-style loop evaluated with IOHprofiler tools 
& Provides a more systematic benchmarking-oriented framework \\ \hline

Multitask GP \cite{zhang2021multitask} 
& Generative hyper-heuristics 
& Evolves heuristics across related tasks using GP and multitask learning 
& Connects LLM-era work to earlier program-evolution ideas \\ \hline

Multiobjective GP \cite{zhang2022multitask} 
& Multi-task heuristic generation 
& Uses task-aware crossover and multi-objective search for scheduling heuristics 
& Shows structured knowledge transfer in heuristic evolution \\ \hline

LEMABE \cite{dashti2022lemabe} 
& Learnable evolution 
& Combines learning-guided evolution with analogy-based estimation 
& Illustrates the broader principle of intelligent variation \\ \hline

\texttt{OptiPattern} \cite{sartori2024metaheuristics} 
& Pattern-guided evolution 
& Uses LLM-derived structural priors to bias BRKGA decoding 
& Shows how LLMs can inject domain knowledge without replacing EC \\ \hline
\end{tabular}}
\begin{flushleft}
\color{black}\footnotesize \textit{Note:} GP: Genetic programming; BRKGA: Biased random-key GA; SOMA: Self-organizing migrating algorithm; ATA: Auto-tuning assistant.
\end{flushleft}
\end{table}

\section{Future Directions and Open Challenges}\label{sec: Future scopes}

The interaction between EC and LLMs is moving beyond isolated proof-of-concept studies toward more integrated, adaptive, and autonomous systems. Earlier sections of this survey examined the two main directions of influence: EC for improving LLMs and LLMs for improving EC. A natural next step is to study frameworks in which the two components interact continuously, exchange information across iterations, and co-adapt over time. From a computer-science perspective, this shift is important because it reframes LLM-EC integration as a broader problem of \emph{automated algorithm design}, \emph{interactive decision support}, and \emph{hybrid intelligent systems engineering}, rather than as a narrow collection of optimization tricks. These directions emerge not from a single line of work, but from recurring patterns observed across the literature reviewed in Sections~\ref{sec: EAs and LLM} and~\ref{sec: LLM and EAs}.

At the same time, the field remains immature. Many current results are encouraging but highly task-dependent, often demonstrated on limited benchmarks, under inconsistent evaluation protocols, or with significant dependence on prompting choices and proprietary model behavior. For this reason, the main value of this section is not to enumerate speculative applications but to identify the most consequential emerging directions and the major obstacles that must be addressed for LLM-EC systems to become reliable scientific and engineering tools \cite{wu2025evolutionary,wang2025large}.

\subsection{Toward Co-Adaptive and Co-Evolutionary LLM-EC Systems}\label{subsec:co_evolutionary}
\color{black}
One of the most important emerging directions is the development of co-adaptive systems, where LLMs and EC interact through repeated feedback and iterative adjustment rather than one-way assistance. In these architectures, EC optimizes components such as prompts, algorithmic components or hyperparameters, while the LLM simultaneously refines its generation strategies or internal guidance based on evolutionary history. This represents a more robust integration than simple hybridization, as both modules participate in a shared adaptive loop \cite{wu2025evolutionary,wang2025large,morris2024llm}.

Figure \ref{fig:llm-ea-coevolution-colored} illustrates this progression, conceptualizing co-adaptive systems not as an isolated third category, but as an emergent synthesis of the two primary interaction channels, EC for LLM and LLM for EC. By combining these channels into bidirectional feedback loops, the framework enables reflective improvement and iterative refinement of both search and reasoning, effectively shifting the system toward a self-improving design pipeline\cite{wu2025evolutionary,morris2024llm}.
\color{black}

\begin{figure}[ht!]
\centering
\scalebox{0.8}{\begin{tikzpicture}[
    box/.style = {draw, rounded corners, thick, text width=7.5cm, align=center, minimum height=2.8cm, fill=#1!20},
    node distance = 5cm and 2.5cm,
    arrow/.style = {thick, -{Stealth[scale=1.1]}, color=black}
]
\node[box=blue] (ea2llm) {\textbf{EC for LLM Enhancement}\\
 Prompt optimization\\
 Hyperparameter tuning\\
 Architecture search\\
 Interaction-policy refinement};

\node[box=red, right=of ea2llm] (llm2ea) {\textbf{LLM for EC Enhancement}\\
 Heuristic generation\\
 Operator and parameter control\\
 Surrogate reasoning\\
 Pattern-guided evolution};

\coordinate (midpoint) at ($(ea2llm)!0.5!(llm2ea)+(0,3)$);
\node[box=green, below=of midpoint] (synergy) {\textbf{Co-Adaptive LLM-EC Systems}\\
 Bidirectional feedback loops\\
 Reflective improvement\\
 Multi-stage algorithm design\\
 Iterative refinement of search and reasoning};

\draw[arrow, line width=2pt] (ea2llm) |- (synergy);
\draw[arrow, line width=2pt] (llm2ea) |- (synergy);
\draw[arrow, line width=2pt] (synergy) -| (ea2llm);
\draw[arrow, line width=2pt] (synergy) -| (llm2ea);
\end{tikzpicture}}
\caption{Conceptual view of the three increasingly integrated modes of interaction between EC and LLMs: EC for LLM enhancement, LLM for EC enhancement, and co-adaptive systems in which both components iteratively influence each other.}
\label{fig:llm-ea-coevolution-colored}
\end{figure}

{ \color{black}In this review, the distinction between \emph{co-adaptive} and \emph{co-evolutionary} systems is important. We use \emph{co-adaptive} in a broad sense to refer to iterative LLM-EC frameworks in which the two sides influence one another through feedback, repeated revision, or staged mutual adjustment, even when only one component is directly optimized at a given step. By contrast, \emph{co-evolutionary} is used in a narrower sense for settings in which both components constitute explicitly adapting entities, and the update of each side depends on the behavior, outputs, or performance of the other \cite{potter2000cooperative,eiben2015introduction}. Under this distinction, many current LLM-EC frameworks are more accurately described as co-adaptive rather than fully co-evolutionary.}

Conceptually, co-adaptive LLM-EC systems can operate at multiple levels. At the lowest level, the LLM may provide local guidance while EC performs a global search. At an intermediate level, the LLM may revise operators, heuristics, or candidate programs based on population statistics and evaluation feedback. At the highest level, the entire system may behave as a self-improving design pipeline in which prompts, heuristics, and algorithmic structures are iteratively generated, evaluated, and refined. Existing frameworks such as PhaseEvo, PromptBreeder, ReEvo, LLaMEA, and related systems suggest the feasibility of such loops, but current evidence remains preliminary and does not yet establish robust long-term co-adaptive behavior across diverse tasks or evaluation settings \cite{van2025llamea,fernando2023promptbreeder,cui2024phaseevounifiedincontextprompt,ye2024reevo}.

{\color{black}A key open question is how to distinguish genuine co-evolutionary behavior from looser forms of tool coupling or one-sided adaptation.} In many current studies, only one component truly adapts, while the other remains fixed. For future work, a more rigorous notion of co-adaptation will be useful, one that specifies which parts of the LLM-EC pipeline evolve, how feedback is represented, and how mutual improvement is measured. Clarifying this distinction will help prevent conceptual overstatement and improve comparability across studies.

\subsection{Future Directions in Algorithm Discovery}\label{subsec:future_directions_algos}

A second major direction is the use of LLMs for discovering new optimization algorithms. This theme has already appeared in recent work on LLM-generated heuristics, reflective search, and language-mediated evolutionary operators, but it remains methodologically underdeveloped \cite{pluhacek2024using,pluhacek2023leveraging,van2025llamea,liu2024evolution,ye2024reevo,zhong2024leveraging}. The central challenge is that ``novel algorithm design'' is often claimed more easily than it is established. In review terms, true algorithmic novelty should not be assessed only by narrative originality or biological metaphor, but by whether a method introduces meaningful search mechanisms, demonstrates robust empirical advantage, and generalizes beyond a narrow benchmark set. 

{\color{black}In the present survey, algorithm discovery is treated not as a separate taxonomic class parallel to the Section~\ref{sec: LLM and EAs} categories, but as a broader future-oriented perspective under which several of those categories may evolve. In particular, both hyper-heuristic generation and LLM-generated metaheuristics can be viewed as contributing to algorithm discovery, but they differ in the level at which the generated artifact operates and in the strength of evidence currently available for genuine novelty.}

{\color{black}More explicitly, algorithmic novelty in LLM-assisted design should be evaluated at several levels rather than treated as a binary label. A generated method may appear novel at the narrative level because it uses a new metaphor, explanation, or naming convention, while remaining close to familiar templates at the mechanism level. For this reason, stronger novelty claims should be grounded in at least three complementary forms of evidence: (i) \emph{mechanism-level novelty}, meaning that the method introduces a meaningfully distinct search operator, population-update logic, selection rule, memory structure, or interaction pattern rather than a superficial restatement of a known family; (ii) \emph{empirical novelty}, meaning that the method demonstrates robust value under matched-budget evaluation against strong baseline algorithms across held-out tasks or benchmark families; and (iii) \emph{traceability and contamination-aware assessment}, meaning that the generated code or algorithmic structure is compared against known template families and interpreted with caution when training-data leakage, public-code overlap, or benchmark memorization may plausibly explain the output.}

For this reason, future research on LLM-assisted algorithm discovery should distinguish more clearly between at least two cases. The first is \emph{improvement of existing algorithms}, where the LLM modifies a known baseline by proposing new operators, structures, or hybrid combinations \cite{pluhacek2024using,pluhacek2023leveraging,pluhacek2023investigating}. The second is \emph{de novo algorithm generation}, where the system attempts to derive a search procedure from broader optimization principles without relying mainly on rearrangements of familiar templates such as GA, PSO, or DE \cite{van2025llamea,zhong2024leveraging}. Current evidence suggests that the first case is considerably more mature than the second. { \color{black}In both cases, claims of novelty should be supported not only by performance results, but also by explicit comparison to known algorithmic families, clear identification of the mechanism that is genuinely new, and discussion of whether the generated structure may reflect recombination of publicly available templates.}

A useful design framework for future work is therefore iterative and diagnostic rather than purely generative. An LLM may propose candidate algorithmic changes, but those candidates should then be tested against well-chosen baselines, analyzed for actual novelty at the mechanism level, and refined using structured feedback \cite{morris2024llm,zhao2022autoopt}. In this sense, human-LLM collaboration remains important. Rather than treating the human as an external evaluator only, future frameworks may benefit from explicitly incorporating expert critique, failure analysis, and interpretability checks into the algorithm-discovery loop. \textcolor{black}{This iterative structure is summarized in Figure~\ref{fig: flowchart}. The loop begins with a prompting stage that takes as input both the optimization task and design objective and any accumulated human critique from prior iterations, and uses these to instruct the LLM to generate a candidate algorithmic structure and accompanying code. This candidate is then passed to an evaluation stage that assesses novelty, robustness, and performance against strong baselines, following the mechanism-level, empirical, and traceability criteria discussed above. The outcome of this evaluation is twofold: candidates that do not meet these criteria are routed back into the loop as critique that informs the next round of prompting, while candidates that pass are advanced to the final stage as a validated algorithmic contribution. This depiction emphasizes that algorithm discovery is not a single generative step but a cycle in which human judgment and empirical scrutiny are applied repeatedly before any claim of novelty is accepted.}
\begin{figure}[ht!]
    \centering
 \resizebox{0.8\linewidth}{!}{\begin{tikzpicture}[node distance=4.5cm, auto]
        \tikzstyle{block} = [rectangle, draw, text width=8em, text centered, rounded corners, minimum height=2.5em]
        \tikzstyle{arrow}=[{thick, -{Stealth[scale=1.1]}, color=black}]

        \node(PT) [fill=orange!50, block]  {Prompting};
        \node(LM) [fill=orange!40, block, right of = PT]  {LLM};
        \node(NA) [fill=orange!30, block, right of = LM, xshift=2]  {Generate candidate algorithm and code};
        \node(PF) [fill=orange!20, block, below of = LM, yshift =2.5cm]  
        {Evaluate: novelty, robustness, and benchmarking};
        \node(CT) [fill=orange!50, block, below of = PT, yshift = 1.0cm]  {Human critique and creative refinement};
        \node(PM) [fill=orange!60, block, left of = PT ]  
        {Optimization task and design objective};
        \node(S) [fill=orange!10, block, below of = PF,  yshift = 2.5cm ]  {Validated algorithmic contribution};

        \draw [arrow] (PT) -- (LM) ;
        \draw [arrow] (LM) --(NA) ;
        \draw [arrow] (NA) |- (PF) ;
        \draw [arrow] (CT) -- (PT) ;
        \draw [arrow] (PM) -- (PT) ;
        \draw [arrow] (PF) -- (S) ;
        \draw [arrow] (PF.west) -- (0, -2);
    \end{tikzpicture}}
   \caption{High-level view of an LLM-assisted algorithm-discovery loop in which candidate designs are generated, critically evaluated, and refined through feedback before being treated as validated contributions.}
    \label{fig: flowchart}
\end{figure}

\subsection{Decision Support, Explainability, and Human-in-the-Loop Systems}

Another promising direction lies in decision support. Although many EC methods are powerful, they often remain difficult for non-experts to formulate, configure, and interpret. LLMs offer a natural interface layer through which users may describe problems in ordinary language, receive suggestions about optimization formulation, obtain recommendations for suitable algorithms, and request explanations of why particular search decisions were made \cite{ramamonjison2023nl4opt,bertsimas2024robust}. In this way, LLMs could substantially broaden access to optimization technology beyond specialist communities.

Within this theme, explainability is especially important. Several current LLM-enhanced EC frameworks already hint at more interpretable evolutionary workflows. Some systems generate natural-language ``thoughts'' alongside executable heuristics, whereas others translate surrogate decisions into textual inferences or use structured prompts to expose how problem patterns influence search bias. These are not yet fully explainable EC frameworks, but they point toward an important design principle: LLMs can make evolutionary search more communicable, self-documenting, and interactive \cite{arrieta2020explainable,gupta2025enhancing,zhao2024explainability,miller2019explanation}.

Human-in-the-loop optimization is likely to be particularly valuable in this setting \cite{amershi2019guidelines}. Rather than aiming for full autonomy in every application, future platforms may combine human problem knowledge, LLM-based reasoning, and EC-based search within a collaborative loop. In practical terms, this could support problem formulation, algorithm recommendation, adaptive hyperparameter tuning, and post-hoc explanation of search behavior \cite{martinek2024large,ramamonjison2023nl4opt,bertsimas2024robust}. This direction is also consistent with earlier examples in which LLMs contribute natural-language rationale, surrogate judgments, or adaptive recommendations without fully replacing the evolutionary search process. The key issue is therefore not whether such systems are possible in principle, but how to design them so that human intervention improves reliability rather than merely masking methodological weaknesses \cite{boussioux2024}.

\textcolor{black}{Finally, the automation of code generation and execution in LLM-EC systems introduces distinct safety risks. Unlike conventional LLM applications, these loops compile and execute unreviewed code across large populations, creating a structural bottleneck where human auditing of every candidate is impractical \cite{gu2026overlooked}. To mitigate risks such as execution-environment compromise, reward hacking, and unsafe exploration, future work must transition from ad-hoc implementation to a standard of safety by design \cite{nilsen2023reward}. This includes: (i) strict sandboxing and resource-bounded execution to isolate generated code; (ii) explicit auditing of fitness functions to detect harness-exploiting or degenerate behavior; and (iii) staged autonomy, where the scale of unsupervised execution is gated by prior reliability checks. Reporting these safety practices should be treated as a necessary component of methodological transparency, complementary to the benchmark reporting standards discussed in Section 4.4.}

\subsection{Benchmarking, Evaluation, and Reproducibility}
Benchmarking remains one of the most urgent methodological issues in LLM-EC research. Many current studies evaluate generated heuristics or hybrid pipelines on narrow benchmark sets, with limited attention to cross-instance generalization, computational budget, prompt sensitivity, or statistical robustness. This creates a serious risk of overinterpreting isolated successes. Recent work has already shown that LLM-generated or LLM-guided heuristics can appear impressive under favorable conditions while failing to generalize under broader or more rigorous evaluation \cite{sim2025beyond,hajari2024searching}. In this setting, benchmarking is not merely a reporting convention, but a core part of establishing whether an apparently novel LLM-EC system represents genuine methodological progress.

Future benchmarking protocols for LLM-EC systems should therefore go beyond single-best-score comparisons and define clearer minimum expectations for reporting, robustness testing, and claim validity. At minimum, evaluation should include: (i) diverse benchmark suites spanning continuous, combinatorial, and algorithm-design tasks; (ii) multiple independent runs with variance reporting; (iii) ablation studies that separate the contribution of the LLM from that of the evolutionary search itself; (iv) reporting of prompt templates, seeds, model versions, and compute budgets; and (v) robustness tests under distribution shift, prompt perturbation, or alternative evaluators. These practices are especially important because both EC and LLM inference are stochastic, making reproducibility substantially harder than in conventional deterministic pipelines \cite{auger2024challenges,bartz2020benchmarking}.

{\color{black}More concretely, a minimally credible evaluation protocol for LLM-EC systems should report at least the following: (i) the intended evaluation regime, including whether the claim concerns broad algorithm discovery or problem-specific design; (ii) the benchmark scope, including the number and diversity of tasks, instances, or suites used for assessment; (iii) multiple independent runs with seed disclosure and variance reporting; (iv) ablation studies that separate the contribution of the LLM, the evolutionary search component, and any auxiliary tuning or reflection mechanism; (v) comparisons against strong manual and automated baselines under matched or clearly disclosed computational budgets; and (vi) full disclosure of prompts, model versions, API settings, inference budgets, and stopping criteria.

\color{black}The evaluation design should also reflect the type of claim being made. For general-purpose algorithm discovery, stronger evidence should come from diverse benchmark families, held-out problem classes, and robustness across prompt formulations, model versions, or evaluator choices. By contrast, for problem-specific algorithm design, narrower evaluation can be acceptable, but studies should then report held-out instances, adaptation budget, reliability under deployment constraints, and clear failure cases rather than presenting narrow benchmark success as broad methodological progress. In both cases, contamination-aware testing and mechanism-level novelty assessment should accompany performance claims whenever the generated outputs resemble known algorithm families or publicly available code patterns \cite{dong2024generalization}.  \color{black}Accordingly, broad benchmark coverage should be interpreted as a requirement for claims of general algorithmic scope, not as a universal requirement for every LLM-EC design setting.

\color{black}This distinction also reflects a broader change in the economics of algorithm development. In traditional EC research, proposing a new algorithm often required substantial manual effort, which naturally encouraged validation on broad benchmark suites to justify that cost. In semi-automated LLM-assisted workflows, however, it may become far more practical to generate specialized algorithms for narrower problem classes or instance families. This shift does not reduce the need for rigorous evaluation, but it does change what kind of evaluation is most appropriate: broad benchmark coverage is essential when claiming general methodological contribution, whereas tailored evaluation may be more appropriate when the goal is reliable and cost-effective performance on a specific target problem class. Accordingly, broad benchmark coverage should be interpreted as a requirement for claims of general algorithmic scope, not as a universal requirement for every LLM-EC design setting.}

{ \color{black}At the same time, direct quantitative comparison across the reviewed literature remains difficult because studies differ substantially in tasks, benchmark suites, evaluation budgets, model access, and reporting protocols. For this reason, Table~\ref{tab:comparative_synthesis} does not attempt a formal meta-analysis of incompatible raw metrics. Instead, it provides a structured comparative synthesis of the major interaction modes, based on reported performance evidence, computational burden, reproducibility, and generalization, using qualitative ordinal judgments grounded in the breadth and rigor of the published evaluations.}

\begin{table}[h!]
\centering
\caption{\color{black}Comparative synthesis of major LLM-EC interaction modes (IntM) in terms of reported performance evidence (RPE), computational burden (CompB), reproducibility (Rep), and generalization (Gen). The ratings are qualitative and ordinal rather than meta-analytic because the reviewed studies differ substantially in tasks, benchmarks, model access, evaluation budgets, and reporting protocols.}
\label{tab:comparative_synthesis}
\resizebox{1\textwidth}{!}{%
\begin{tabular}{p{3.8cm}p{4cm}p{2.4cm}p{2.6cm}p{2.6cm}p{8cm}}
\toprule
\textbf{IntM} & \textbf{RPE} & \textbf{CompB} & \textbf{Rep} & \textbf{Gen} & \textbf{Interpretive Summary} \\
\midrule
Hard-prompt optimization & Moderate to Strong & Moderate & Moderate & Limited to Moderate & Most mature EC-for-LLM direction; multiple studies report useful gains, but evidence remains concentrated in relatively narrow task families and benchmark settings. \\
\midrule
Soft-prompt evolution & Limited & High & Limited & Limited & Conceptually promising, but still immature from an evolutionary perspective, with underdeveloped operators and limited comparative validation. \\
\midrule
EC-based hyperparameter tuning & Limited to Moderate & Moderate to High & Limited & Limited & Useful in heterogeneous, black-box, or multi-objective settings, but evidence remains fragmented across different optimization targets and tasks. \\
\midrule
Evolutionary NAS for LLM-related systems & Limited to Moderate & High & Limited & Limited & Methodologically relevant, but strongly constrained by evaluation cost and relatively weak large-scale comparative evidence. \\
\midrule
LLM-assisted heuristic/hyper-heuristic generation & Moderate & Moderate & Limited & Limited to Moderate & Stronger conceptual and methodological evidence than broad performance evidence; current results are more convincing for feasibility than for consistent superiority over strong classical baselines. \\
\midrule
LLM-assisted surrogate reasoning and adaptive control & Limited to Moderate & Moderate & Limited & Limited & Expands the role of LLMs within EC pipelines, but remains exploratory and not yet broadly validated against established surrogate or control methods. \\
\midrule
LLM-generated metaheuristics and language-mediated search & Limited & Moderate & Limited & Limited & Conceptually significant but empirically unsettled; claims of novelty and broad performance advantage remain much easier to assert than to validate. \\
\midrule
Co-adaptive LLM-EC systems & Very Limited & High & Very Limited & Very Limited & Important emerging direction, but still dominated by proof-of-concept studies with limited long-horizon, cross-task, or contamination-aware validation. \\
\bottomrule
\end{tabular}}
\end{table}

{\color{black}Building on the minimally credible evaluation protocol outlined above, Fig.~\ref{fig:llm_ec_evaluation_framework_hex} consolidates these components into a single visual summary highlighting the core dimensions of credible LLM-EC evaluation, especially for settings that claim broad algorithmic scope or general-purpose methodological contribution. In narrower problem-specific design settings, these dimensions remain relevant, but their implementation should be adapted to the target problem class, deployment constraints, and intended use context.}

\begin{figure}[h!]
\centering
\begin{tikzpicture}[
    >=Stealth,
    hex/.style={draw, regular polygon, regular polygon sides=6, minimum width=4.2cm, align=center, font=\small, fill=purple!20},
    box/.style={draw, rounded corners, text width=3.8cm, minimum height=1.15cm, align=center, font=\small, fill=white},
    link/.style={thick, -{Stealth[length=2.5mm]}}
]

\node[hex] (core) at (0,0) {\textbf{Credible}\\\textbf{LLM-EC}\\\textbf{Evaluation}};

\node[box,fill=orange!30] (b1) at (0,4.0) {\textbf{Benchmark breadth}\\Continuous, combinatorial, and algorithm-design tasks};

\node[box,fill=orange!25] (b2) at (4.4,2.2) {\textbf{Multiple runs /variance}\\Independent trials and statistical robustness};

\node[box,fill=orange!20] (b3) at (4.4,-2.2) {\textbf{Ablations}\\Separate LLM effects from EC effects};

\node[box,fill=orange!15] (b4) at (0,-4.0) {\textbf{Compute-budget reporting}\\Inference cost, search cost, and resource use};

\node[box,fill=orange!10] (b5) at (-4.4,-2.2) {\textbf{Prompt/model disclosure}\\Prompts, seeds, versions, and settings};

\node[box,fill=orange!5] (b6) at (-4.4,2.2) {\textbf{Contamination/leakage checks}\\Memorization, source overlap, and originality};

\draw[link] (b1) -- (core);
\draw[link] (b2) -- (core);
\draw[link] (b3) -- (core);
\draw[link] (b4) -- (core);
\draw[link] (b5) -- (core);
\draw[link] (b6) -- (core);


\end{tikzpicture}
\caption{Layered evaluation framework for credible LLM-EC research. Reliable assessment depends not only on reported performance but also on benchmark breadth, repeated evaluation, ablation studies, compute-budget transparency, prompt and model disclosure, and contamination-aware testing.}
\label{fig:llm_ec_evaluation_framework_hex}
\end{figure}










\color{black}
\subsection{Memorization and Contamination in LLM-Generated Algorithms}
Among the credible LLM-EC evaluation dimensions summarized in Fig.~\ref{fig:llm_ec_evaluation_framework_hex}, contamination and memorization deserve separate and more detailed treatment given their methodological importance and their relevance to several directions discussed above \cite{dong2024generalization}.
Since many powerful LLMs are trained on large and largely undisclosed corpora, it is often difficult to determine whether a strong algorithmic output reflects genuine reasoning (novelty) over the stated problem or the recovery of patterns already present in training data. This concern is distinct from, but closely related to, the traceability criterion introduced in Section \ref{subsec:future_directions_algos}, which asks whether a generated structure can be matched to known algorithm families, whereas contamination asks whether the underlying training process had direct or indirect access to the specific benchmark, dataset, or solution being evaluated \cite{dong2024generalization}.

This issue is particularly serious when the target tasks involve well-known benchmark problems, canonical algorithm templates, or widely available source code, since in such cases an impressive result may partly reflect memorized implementations, common template structures, or training-data leakage rather than genuine algorithmic discovery. The risk is compounded in LLM-generated metaheuristics and hyper-heuristic frameworks (Sections \ref{subsubsec: LLM and HHs}-\ref{subsec: LLM and MHs}), where the LLM is directly responsible for producing executable code. In such settings, a method that performs well may do so because it closely reproduces a publicly documented routine for that specific benchmark, rather than because the search process discovered a mechanism that generalizes beyond it. Co-adaptive systems (Section \ref{subsec:co_evolutionary}) and de novo algorithm-discovery claims (Section \ref{subsec:future_directions_algos}) are especially exposed to this risk, since both rely on extended generation and refinement loops in which contaminated outputs can be reinforced across iterations rather than identified and discarded.

Mitigating this risk requires evaluation practices that go beyond reporting performance comparison. As a baseline practice, future studies should: (i) disclose, where possible, whether the target benchmark or closely related instances are likely to appear in public training corpora; (ii) compare generated algorithms against known template families to assess whether claimed novelty is mechanism-level or merely a restatement of memorized structure; (iii) test on held-out, recently introduced, or synthetically constructed problem instances that are unlikely to have been seen during pretraining; and (iv) treat unusually strong zero-shot performance on canonical benchmarks with particular caution rather than as default evidence of genuine discovery. These practices complement, rather than replace, the broader evaluation dimensions summarized in Fig.~\ref{fig:llm_ec_evaluation_framework_hex}, and should be considered a necessary component of credible reporting whenever an LLM is used to generate or substantially modify executable algorithmic code \cite{dong2024generalization}.
\color{black}

\subsection{Core Open Challenges}
The most important open challenges in the field can be grouped into five categories.

\textbf{First, computational complexity and scalability.} Hybrid LLM-EC systems are often expensive because they combine repeated evolutionary evaluation with costly model inference or generation. This limits feasible population sizes, benchmark breadth, and accessibility for many research groups. \textcolor{black}{More efficient surrogates, selective evaluation, and lower-cost interaction protocols will be essential. A concrete future step will be reporting a compute-normalized comparison (e.g., performance per LLM-inference call or per token budget) alongside raw performance, so that gains are not confounded with sheer evaluation expenditure, as argued for cost-controlled agent evaluation more broadly \cite{kapoor2024ai}. Existing EC benchmarking infrastructure such as IOHprofiler already logs granular per-evaluation trajectories \cite{doerr2018iohprofiler, de2024iohexperimenter, wang2022iohanalyzer}, and extending this logging to also capture LLM-inference cost would allow such compute-normalized comparisons to be adopted with minimal additional changes.}

\textbf{Second, weak theoretical grounding.} 
\textcolor{black}{While classical evolutionary computation has developed rigorous theoretical foundations for runtime analysis, convergence, and fitness-landscape analysis \cite{hao2024large,eiben2015introduction}, comparable theory for LLM-EC systems remains unexplored in the current literature. Existing approaches have primarily been validated through empirical evaluation, leaving fundamental questions about how language-guided operators influence convergence, search-space exploration, diversity preservation, and optimization performance across problem classes. Future work should develop formal models of LLM-generated variation operators, characterize their effects on symbolic and executable search spaces, derive theoretical guarantees on convergence and runtime where possible, and identify the problem characteristics under which semantic guidance improves or hinders evolutionary search.}

\textbf{Third, fitness design and semantic evaluation.} Hybrid systems often optimize outputs that are not easily captured by a single numerical metric. When LLMs generate prompts, heuristics, code, or explanations, evaluation may need to account for correctness, novelty, semantic validity, robustness, and cost simultaneously. Designing principled multi-criteria fitness functions remains an open problem.

\textbf{Fourth, interpretability and causal understanding:} LLMs can generate plausible algorithmic variations, but it remains unclear whether they capture the causal mechanisms that make certain search strategies effective. This raises a deeper scientific question: are LLMs discovering algorithmic principles, or mainly recombining patterns seen during training? 
\textcolor{black}{Addressing this question requires evaluation beyond solution quality. Future work should develop diagnostic benchmarks that distinguish genuine algorithmic abstraction from memorization, including counterfactual optimization problems, intervention-based evaluations of search dynamics, and tasks requiring symbolic explanation alongside empirical validation \cite{romera2024mathematical,geiger2025causal, xu2024faithful}. Integrating these benchmarks with mechanistic interpretability and symbolic reasoning techniques \cite{brinkmann2024mechanistic,somvanshi2026bridging} could provide stronger evidence about whether LLMs acquire transferable algorithmic principles or mainly exploit statistical regularities in their training data.}

\textbf{Fifth, system-level integration.} Distributed, federated, and long-horizon hybrid systems raise additional issues of communication, memory, catastrophic forgetting, and coordination across evolving components. These settings remain largely unexplored, yet they are likely to become increasingly important as LLM-EC frameworks grow in scope and autonomy.

Taken together, these challenges suggest that the next phase of research should emphasize methodological discipline as much as algorithmic creativity. The field has already shown that LLMs and EC can be combined in interesting ways; the harder, more important task now is to determine how to make these combinations robust, interpretable, reproducible, and scientifically credible.

\subsection{Limitations of the Present Review}
{\color{black}Like the literature it surveys, the present review has several limitations. First, this article is structured as a critical narrative review rather than a fully systematic review or meta-analysis. Although the search and screening process has been made more transparent, the review does not claim exhaustive coverage of every relevant publication in this rapidly expanding area. Second, the field itself is highly dynamic and partly shaped by recent conference papers, workshop papers, and preprints, which means that the maturity, reproducibility, and archival stability of the underlying evidence remain uneven across topics.

Third, some of the methodological categories used in this survey are analytically useful but not fully disjoint. Several recent LLM-EC frameworks combine multiple roles within the same pipeline, including heuristic generation, reflective revision, surrogate assistance, and algorithm-level design. For this reason, the proposed taxonomy should be interpreted as a structured organizing framework rather than as a final or universally fixed classification of the field. Fourth, the empirical evidence reviewed in the paper is heterogeneous in both quality and scope. In several areas, especially co-adaptive systems and LLM-generated algorithm design, the literature is still dominated by proof-of-concept studies, limited benchmark sets, and inconsistent reporting practices.

These limitations do not reduce the importance of the reviewed direction, but they do shape how the present survey should be interpreted. The paper is intended to provide a transparent, critical, and conceptually organized synthesis of the most relevant developments to date, while also making clear where the current evidence base remains incomplete, unstable, or methodologically underdeveloped.}

\section{Conclusion}\label{sec: conclusion}
This survey reviewed the emerging interplay between evolutionary computation (EC) and large language models (LLMs) from a bidirectional perspective. On one side, EC enhances LLM-centered systems through prompt optimization, hyperparameter tuning, and architecture search. On the other, LLMs enhance EC through heuristic generation, surrogate reasoning, adaptive control, and automated algorithm design. Across both directions, a broader pattern emerges: LLMs contribute abstraction, semantic reasoning, and generative flexibility, whereas EC contributes structured search, diversity maintenance, and robustness in difficult optimization spaces. The survey also highlighted a more ambitious frontier in which LLMs and EC interact through co-adaptive loops. This direction is especially significant because it shifts the focus from isolated method enhancement toward automated algorithm design and hybrid intelligent systems. However, the current literature remains uneven. Many results are still proof-of-concept and are supported by limited benchmarks, inconsistent evaluation, or insufficient attention to reproducibility and generalization. As a result, the field should be regarded as promising but methodologically unsettled. {\color{black}The present review should therefore be read with appropriate caution: it offers a structured and critical synthesis of the current literature, but not an exhaustive or final account of a rapidly evolving field.}
Several priorities emerge for future research. First, stronger benchmarking and reporting standards are needed to support fair comparison and cumulative progress. Second, hybrid systems require firmer theoretical grounding, especially regarding search dynamics, causal understanding, and the interaction between language-based reasoning and evolutionary search. Third, future work should place greater emphasis on interpretability, human-AI collaboration, and practically useful decision-support frameworks rather than on autonomous generation alone. Finally, scalability, computational cost, and contamination-aware evaluation will remain central barriers to broader adoption.
Overall, the significance of LLM-EC integration lies not only in improved performance on specific tasks, but also in its potential to reshape how optimization algorithms are designed, adapted, and deployed. By connecting linguistic intelligence with evolutionary search, this emerging area opens a path toward more flexible, interactive, and self-improving computational systems. Whether that potential can be realized will depend less on isolated benchmark wins than on the development of rigorous, transparent, and conceptually sound research practices. {\color{black}More broadly, the reviewed literature points less to a single application taxonomy than to recurring interaction roles, including prompt adaptation, heuristic generation, surrogate reasoning, operator guidance, and algorithm-level design support. Taken together, these features distinguish the present survey from adjacent reviews that focus mainly on one interaction direction, one methodological family, or broader LLM-based optimization without an EC-aware critical synthesis.}

 \section*{Acknowledgment}
 This work was supported by Dr. B. R. Ambedkar National Institute of Technology Jalandhar; the Anusandhan National Research Foundation, Government of India (Award No. MTR/2021/000503), the Australian Researcher Cooperation Hub through the Australia-India Women Researchers' Exchange Program; and the Spanish Ministry of Economy and Competitiveness through the Ramón y Cajal Research Grant (Award No. RYC2023-045020-I).
 
\section*{Compliance with ethical standards}
	\noindent \textbf{Conflict of interest:} All the authors declare that they have no conflict of interest.
	
	\noindent \textbf{Ethical approval:} This article does not contain any studies with human participants or animals performed by any of the authors.

\appendix

\section{Supplementary Technical Details for LLM-Enhanced EC Frameworks}\label{secapp:appA}

This appendix collects implementation-oriented details that support the discussion in Sections~\ref{sec: EAs and LLM} and \ref{sec: LLM and EAs} but are not essential to the main survey narrative. These materials are included to preserve methodological transparency while keeping the main paper focused on conceptual synthesis, comparative analysis, and open research challenges.

\subsection{Supplementary Tables for EC-Based LLM Enhancement}\label{secapp:ec_llm_tables}
This subsection collects detailed comparative tables moved from Section~\ref{sec: EAs and LLM} to keep the main text focused on synthesis rather than inventory-style presentation.

\subsubsection{Supplementary Summary of Representative Prompt-Optimization Frameworks}\label{app:prompt_tools_supplement}
{\color{black}This supplementary subsection provides additional comparative detail for the three representative prompt-optimization frameworks discussed in Section~\ref{subsec:hard_prompting}. As explained in the main text, EvoPrompt, PhaseEvo, and GAAPO were selected for focused comparison because they illustrate three distinct methodological design strategies in evolutionary prompt optimization: direct LLM-based evolutionary operators, structured multi-phase optimization, and hybrid orchestration of multiple prompt-generation strategies. %
}
\subsubsection{Supplementary Synthesis of Hyperparameter-Tuning Directions}\label{app:hpo_supplement}
{\color{black}This supplementary subsection reorganizes the hyperparameter-tuning literature discussed in Section~\ref{subsec:hyperparameter} into a more synthetic category-based view. Because the main text emphasizes conceptual interpretation rather than inventory-style coverage, Table~\ref{tab:ea_llm_hpo_app} collects representative hyperparameter types, corresponding evolutionary techniques, and typical example settings in a compact comparative format. The goal is not to claim a large or fully mature body of evidence, but rather to show the main directions in which EC has been used for hyperparameter optimization in LLM-centered pipelines.}

\begin{table}[h!]
\caption{Supplementary comparative synthesis of EA applications in LLM hyperparameter optimization.}
\label{tab:ea_llm_hpo_app}
\centering
\resizebox{1\linewidth}{!}{\begin{tabular}{p{7cm}p{7cm}p{7cm}}
\toprule
 \textbf{Targeted Hyperparameters} & \textbf{Evolutionary Techniques Employed} & \textbf{Case Studies and Examples} \\
\midrule
\textit{Architectural Hyperparameters:} (i) number of Transformer layers, (ii) hidden state dimension, (iii) number of attention heads, and (iv) feed-forward network intermediate size
&
\textit{Custom EC:} (i) selection via performance ranking, (ii) mutation of hyperparameters, and (iii) architecture exploration
&
\textit{AutoTinyBERT and related architecture-oriented studies}
\\ \hline
\textit{Model-Merging Hyperparameters:} (i) parameters for TIES-Merging + DARE and (ii) weight combination strategies
&
\textit{Evolution Strategies/CMA-ES}
&
\textit{Evolutionary model merging for foundation models}
\\ \hline
\textit{EA-Specific Hyperparameters:} (i) step-size in ES, (ii) temperature in LMEA, and (iii) self-adaptation mechanisms
&
\textit{ES, CMA-ES, and LLM-guided adaptation}
&
\textit{LLM-guided step-size control and hybrid LLM-EA tuning}
\\ \hline
\textit{General ML Hyperparameters:} (i) learning rate, batch size, dropout, and (ii) other training controls
&
\textit{GA and PSO}
&
\textit{General autonomous HPO settings}
\\
\bottomrule
\end{tabular}}
\begin{flushleft}
\color{black}\footnotesize \textit{Note:} EA: Evolutionary algorithm; HPO: Hyperparameter optimization; CMA-ES: Covariance matrix adaptation evolution strategy; ES: evolution strategy; GA: Genetic algorithm; PSO: Particle swarm optimization.
\end{flushleft}
\end{table}

\subsubsection{Supplementary Summary of Evolutionary NAS Studies}\label{app:nas_supplement}
{ \color{black}This subsection summarizes representative evolutionary NAS studies discussed in Section~\ref{subsec:NAS}. It is included here to document the broader study landscape without interrupting the main-text emphasis on methodological synthesis, comparative interpretation, and evidence quality (Table \ref{tab:evolutionary_nas_app}).}

\begin{table}[h!]
\caption{Supplementary summary of evolutionary approaches for neural architecture search in LLM-related settings.}
\label{tab:evolutionary_nas_app}
\centering
\resizebox{1\linewidth}{!}{\begin{tabular}{p{4cm}p{4cm}p{4cm}p{4cm}p{5cm}}
\toprule
\textbf{Study/Method} & \textbf{EA/Search Technique} & \textbf{Optimized Aspects} & \textbf{Objectives/Metrics} & \textbf{Key Finding/Contribution} \\
\midrule
AutoBERT-Zero & Custom EA & BERT backbone structure & Performance & Evolved backbone structures using EA-based NAS. \\
DistilBERT NAS & NSGA-II (MOEA) & Attention heads, FFN design, encoder blocks & QA performance vs. model size & Applied MOEA-based NAS under resource constraints. \\
LLMatic & QD (MAP-Elites) + LLM & Code-level architectures & Accuracy + diversity & Dual-archive search with LLM-guided variation. \\
EvoPrompting & LM operator + evolutionary search & Code-level GNN design & Performance + diversity & Uses language-model-guided search for architecture generation. \\
LiteTransformerSearch & MOEA & Decoder-layer hyperparameters & Perplexity vs. latency vs. memory & Training-free MOEA-NAS for efficient GPT-style models. \\
SuperShaper~\cite{ganesan2021supershaper} & EA & Hidden dimensions & Task-agnostic pre-training & Searches hidden dimensions for efficient BERT variants. \\
Klein et al.~\cite{klein2024structural} & MOEA & Subnetwork structures/pruning & Performance vs. model size & Multi-objective structural pruning via NAS. \\
Choong et al.~\cite{choong2023jack} & MO-MFEA & Model configurations & Multi-task performance vs. size & Multi-task evolutionary search for specialized smaller models. \\
GPT-NAS~\cite{yu2023gpt} & EA + LLM & Network architecture & Performance & Combines LLM generation with evolutionary search. \\
Guided Evolution~\cite{morris2024llm} & EA + LLM & Code-level models & Performance & Uses LLM guidance within an evolutionary improvement loop. \\
\bottomrule
\end{tabular}}
\begin{flushleft}
\color{black}\footnotesize \textit{Note:} NAS: Neural architecture search; EA: Evolutionary algorithm; NSGA-II: Non-dominated sorting GA II; MOEA: Multi-objective EA; QD: Quality-diversity. \\ \textit{Note:} This supplementary table includes both core LLM-related NAS studies and a small number of adjacent code-level architecture-generation studies retained for methodological context.
\end{flushleft}
\end{table}

\subsection{Illustrative Surrogate-Model Workflow}\label{app:surrogate_workflow}

Algorithm~\ref{algo:llm_surrogate} summarizes a generic workflow for using an LLM as a surrogate model within an evolutionary optimization loop. It is included for illustration only; in the main text, surrogate modeling is discussed at a conceptual level.

\begin{algorithm}[h!]
\caption{LLM as Surrogate Model}\label{algo:llm_surrogate}
\begin{algorithmic}[1]
\REQUIRE $X$, $Y$, $U$, $LLM$, $Opt$
\ENSURE $\tilde{Y}$
\STATE $\tilde{Y} \gets \emptyset$
\STATE $X, U \gets \text{Preprocessing}(X, U)$
\FOR{$u \in U$}
    \STATE $prompt \gets \text{GeneratePrompt}(X, Y, u, Opt)$
    \STATE $response \gets \text{Inference}(LLM, prompt)$
    \STATE $y \gets \text{PostProcessing}(response, Opt)$
    \STATE $\tilde{Y} \gets \tilde{Y} \cup y$
\ENDFOR
\end{algorithmic}
\end{algorithm}

\subsection{Example Prompt for Regression-Based Surrogate Reasoning}\label{app:regression_prompt}

Figure~\ref{fig:prompt_regression} provides an example of how historical evaluations and a new candidate can be formatted into a regression-style prompt for an LLM-based surrogate. It is kept in the appendix because it is useful for implementation understanding, but too detailed for the main review narrative.

\begin{figure}[h!]
    \centering
\resizebox{1\linewidth}{!}{\begin{tikzpicture}
    \node[draw, minimum width=2.5cm, fill=blue!30, minimum height=0.8cm]at(5,12) (init) {Prompt for regression};
    \node[draw, fill=green!30, minimum width=2cm, minimum height=0.8cm, below=1.8cm of init] (llm) {Procedure};
    \node[draw, fill=orange!30, minimum width=2.5cm, minimum height=0.8cm, below=1.3cm of llm] (selection) {Historical examples};
    \node[draw, fill=red!30, minimum width=2.5cm, minimum height=0.8cm, below=0.8cm of selection] (crossover) {New evaluation};

    \node[draw, rounded corners, minimum width=6cm, minimum height=2cm, right=1cm of init, align=left, anchor=west] (llm_prompt) {\textbf{\underline{Prompt for LLM}} \\[2pt]
         \small{\textcolor{red!40!black}{Your task is to predict the numerical value of each object based on its attributes. These attributes and their corresponding values are outcomes of a black box function's }}\\
         \small{\textcolor{red!40!black}{operation within its decision space. The target value for each object is determined by a specific mapping from these attributes through the black box function. }}\\
         \small{\textcolor{black!40!black}{Your objective is to infer the underlying relationships and patterns within the black box function using the provided historical data. This task goes beyond simple}}\\
         \small{\textcolor{black!40!black}{ statistical analyses, such as calculating means or variances, and requires understanding the complex interactions between the attributes.}}\\
         \small{Please do not attempt to fit the function using code similar to Python; instead, directly learn and infer the numerical values.}};

   \node[draw, rounded corners, minimum width=10cm, minimum height=1cm, right=2cm of llm, align=left] (llm_prompt1) {\textcolor{orange!40!black}{
        1. Analyze the historical data to uncover how attributes relate to the numerical values.} \\
        \textcolor{orange!40!black}{
        2. Use these insights to predict the numerical value of new objects based on their attributes.} \\
        \textcolor{orange!40!black}{
        3. Respond using JSON format, e.g. \{`Value': `approximation result'\}}};

     \node[draw, rounded corners, minimum width=10cm, minimum height=2cm, right=1.2cm of selection, align=left] (llm_prompt2) {
        \small{\textcolor{red!70}{Features: $⟨0.338, 0.531, ..., 0.363⟩$; Value: 0.41148}}\\
        \small{\textcolor{red!70}{Features: $⟨0.207, 0.598, ..., 0.285⟩$; Value: 0.35745}} \\
        \small{\textcolor{red!70}{...}} \\
        \small{\textcolor{red!70}{Features: $⟨0.629, 0.029, ..., 0.279⟩$; Value: 0.67179}}};
    \node[draw, rounded corners, minimum width=10cm, minimum height=0.5cm, right=1.6cm of crossover, align=left] (llm_prompt3) {\textcolor{black!70}{\small{Features: $⟨0.189, 0.917, ..., 0.443⟩$ }}};

    \node[draw, rounded corners, minimum width=10cm, minimum height=0.5cm, right=1.6cm of llm_prompt2, align=left] {\small{Note: Respond in JSON with the format \{`Value': `approximation result'\} only.}};

     \draw[->, thick,line width=2pt] (init.east) -- ++(0.5,0) |- (llm_prompt.west);
     \draw[->, thick,line width=2pt] (llm.east) -- ++(0.5,0) |- (llm_prompt1.west);
     \draw[->, thick,line width=2pt] (selection.east) -- ++(0.5,0) |- (llm_prompt2.west);
     \draw[->, thick,line width=2pt] (crossover.east) -- ++(0.5,0) |- (llm_prompt3.west);
\end{tikzpicture}}
    \caption{Example prompt and procedure for LLM-assisted regression-based surrogate modeling \cite{hao2024large}.}
    \label{fig:prompt_regression}
\end{figure}

\subsection{Illustrative Prompt Templates for LLM-Based Parameter Tuning}\label{app:tsp_prompts}

Table~\ref{tab:prompts_tsp} lists representative prompt types used for LLM-guided parameter tuning of metaheuristics. These prompt templates are useful for understanding the interaction style between the optimizer and the LLM, but are too detailed for the main text.

\begin{table}[h!]
    \centering
    \caption{Illustrative prompt types used in Ref.~\cite{martinek2024large} for LLM-guided tuning of metaheuristics.}
    \label{tab:prompts_tsp}
    \resizebox{1\linewidth}{!}{\begin{tabular}{cp{18cm}}\hline
      (i)   & Which metaheuristic algorithms would you use to solve a TSP? \\\hline
      (ii)  & I want to solve TSP (defined on 15 cities) with GA. Its \texttt{mealpy} implementation takes the following parameters: parameters. Advise on parameter values. \\\hline
      (iii) & For the TSP task with 15 cities, you previously suggested parameter values for Genetic Algorithm metaheuristic. Parameters and values. I ran the GA algorithm 100 times and got the average global optimum of 0.375 with a standard deviation of 0.03. I also measured the variance of the population at the beginning and last epoch: 9.598 and 5.675, respectively, with std of 0.09 and 0.58. The solution at the last epoch had an average fitness of 0.45 with a standard deviation of 0.37. What changes to the parameters would you suggest to improve performance? Keep the epoch*pop size constant. \\\hline
    \end{tabular}}
\end{table}

\subsection{Detailed Prompt Structure for LMEA}\label{app:lmea_prompt}

Figure~\ref{fig:lmea_prompt} shows a detailed prompt template for LMEA. Since the main text now discusses LMEA at a conceptual level, the explicit prompt structure is moved here to avoid overloading the survey narrative with implementation detail.

\begin{figure}[h!]
    \centering
\resizebox{1\linewidth}{!}{\begin{tikzpicture}
    \node[draw, minimum width=2.5cm, minimum height=0.8cm]at(5,10) (init) {Initialization};
    \node[draw, fill=green!30, minimum width=2cm, minimum height=0.8cm, below=0.4cm of init] (llm) {LLM};
    \node[draw, fill=orange!30, minimum width=2.5cm, minimum height=0.8cm, below=0.3cm of llm] (selection) {Selection};
    \node[draw, fill=orange!30, minimum width=2.5cm, minimum height=0.8cm, below=0.3cm of selection] (crossover) {Crossover};
    \node[draw, fill=orange!30, minimum width=2.5cm, minimum height=0.8cm, below=0.3cm of crossover] (mutation) {Mutation};
    \node[draw, minimum width=2.5cm, minimum height=0.8cm, below=0.4cm of mutation] (termination) {Termination};

    \draw[->,line width=1pt] (init.south) -- (llm.north);
    \draw[->,line width=1pt] (selection.south) -- (crossover.north);
    \draw[->,line width=1pt] (crossover.south) -- (mutation.north);
    \draw[->,line width=1pt] (mutation.south) -- (termination.north);
    \draw[->,line width=1pt] (5,4.9)--(3.5,4.9)--(3.5,9.45)--(5,9.45);
    \node[fill=blue!30, rotate=-90] at (6.7,6.7){\bf Genetic Algorithm};

    \node[draw, rounded corners, minimum width=10cm, minimum height=6cm, right=1.3cm of llm, align=left, anchor=west] (llm_prompt) {\textbf{\underline{Prompt for LLM}} \\[2pt]
        \textbf{\textcolor{green!40!black}{Description of problem and solution properties}} \\
         \small{You are given a list of points with coordinates: \{points\}.}\\
         \small{Your task is to find a trace, with the shortest possible length, that traverses each point exactly once.} \\[6pt]

         \textbf{\textcolor{green!40!black}{In-context examples (population)}} \\
         \small{Below are some previous traces and their lengths, ordered by their lengths:}\\[2pt]
        \texttt{\{trace 1\} \{Length of trace 1\}} \\
        \texttt{\{trace 2\} \{Length of trace 2\}} \\
        \texttt{\{trace 3\} \{Length of trace 3\}} \\
        \texttt{...} \\
        \texttt{\{trace N\} \{Length of trace N\}} \\[6pt]

        \textbf{\textcolor{green!40!black}{Task instructions}} \\
        \small{1. \textcolor{red}{Select} two traces from the above.} \\
        \small{2. \textcolor{red}{Crossover} the two traces from Step 1 and generate a new trace.} \\
        \small{3. \textcolor{red}{Mutate} the trace from Step 2 to generate a new trace.} \\
        \small{4. Repeat the process until \{N\} traces are generated.}\\[6pt]
        \small{Provide traces in XML-like format using \texttt{$<selection>$} and \texttt{$<res>$} tags.}};

     \draw[<->, thick,line width=2pt] (llm.east) -- ++(1,0) |- (llm_prompt.west);
\end{tikzpicture}}
    \caption{Detailed prompt structure used in LMEA for solving TSP-like problems through language-mediated evolutionary search \cite{liu2024large}.}
    \label{fig:lmea_prompt}
\end{figure}

\section{Background Note on Prompt Engineering}\label{secapp:appB}

This appendix note provides a brief background on prompt engineering concepts referenced in Section~\ref{sec: EAs and LLM}. It is intentionally concise, since the present article is a survey on the interaction between LLMs and EC rather than a tutorial on prompting.

\subsection{Hard and Soft Prompts}

In the context of LLMs, a \textit{prompt} is the textual or embedded input sequence that conditions the model to perform a desired task. Prompts are commonly divided into two broad categories \cite{chen2025unleashing,sahoo2024systematic,oppenlaender2024prompting}: \textit{hard prompts}, which are explicit human-readable instructions, and \textit{soft prompts}, which are trainable continuous vectors inserted into the model’s embedding space. Hard prompts are transparent and interpretable, whereas soft prompts offer more flexible and compact task adaptation at the representation level.

This distinction is directly relevant to EC-based optimization. Hard prompts define a discrete textual search space, making them natural targets for token-level mutation and crossover. Soft prompts define a continuous embedding space, which is harder to interpret but opens the door to richer forms of adaptation and hybrid search.

\begin{figure}[h!]
\centering
\resizebox{0.7\linewidth}{!}{
\begin{tikzpicture}[
    root/.style={
        draw,
        rounded corners,
        fill=purple!20,
        text width=6cm,
        minimum height=1.5cm,
        align=center,
        font=\bfseries\large,
        drop shadow
    },
    child/.style={
        draw,
        rounded corners,
        text width=5cm,
        minimum height=1.5cm,
        align=center,
        font=\normalsize,
        inner sep=5pt
    },
    groupbox/.style={
        draw,
        rounded corners,
        thick,
        text width=6cm,
        minimum height=1.2cm,
        align=center,
        font=\bfseries
    },
    arrow/.style={
        -Stealth,
        thick,
        shorten >=2pt,
        shorten <=2pt
    }
]

\node[root] (root) {Prompt in LLMs};

\node[groupbox, fill=blue!20, below=1cm and 1cm of root] (hardbox) {Hard Prompt (Text-based)};
\node[child, fill=blue!15, below=0.5cm of hardbox] (instruction) {\textbf{Instruction}\\\footnotesize Defines task (e.g., ``Summarize this text'')};
\node[child, fill=blue!10, below=0.5cm of instruction] (context) {\textbf{Context/Examples}\\\footnotesize Few-shot examples (input-output pairs)};
\node[child, fill=blue!5, below=0.5cm of context] (query) {\textbf{Input/Query}\\\footnotesize User request or test case};

\node[groupbox, fill=orange!20, below right=1cm and 1cm of root] (softbox) {Soft Prompt (Embedding-based)};
\node[child, fill=orange!15, below=0.5cm of softbox] (vectors) {\textbf{Prompt Vectors}\\\footnotesize Learned continuous embeddings};
\node[child, fill=orange!10, below=.5cm of vectors] (insertion) {\textbf{Insertion Layer}\\\footnotesize Injected into model embedding space};
\node[child, fill=orange!5, below=.5cm of insertion] (adaptation) {\textbf{Task Adaptation}\\\footnotesize Guides model via internal activations};

\draw[arrow] (root.south) -- ++(0,-0.5) -| (hardbox.north);
\draw[arrow] (root.south) -- ++(0,-0.5) -| (softbox.north);
\draw[arrow] (hardbox.south) -- (instruction.north);
\draw[arrow] (instruction.south) -- (context.north);
\draw[arrow] (context.south) -- (query.north);
\draw[arrow] (softbox.south) -- (vectors.north);
\draw[arrow] (vectors.south) -- (insertion.north);
\draw[arrow] (insertion.south) -- (adaptation.north);
\end{tikzpicture}
}
\caption{Conceptual components of hard and soft prompting in LLMs.}
\label{fig:components_llm_prompt}
\end{figure}

\subsection{Prompting as an Optimization Target}

Prompt engineering has evolved from manual trial-and-error toward increasingly systematic optimization. Few-shot prompting, chain-of-thought prompting, zero-shot-CoT, and multi-step prompting strategies have shown that prompt structure can strongly affect downstream performance \cite{ye2023prompt,yao2023more,diao2023active,wang2023msprompt}. These observations motivate the use of EC as an automated search mechanism over prompt representations.

In the context of this survey, the importance of prompt engineering is therefore not purely linguistic. It also provides a concrete example of how evolutionary search can operate over structured, semantically meaningful design spaces, thereby illustrating one of the clearest interfaces between EC and LLMs.

\bibliographystyle{unsrt}
\bibliography{references}
\end{document}